\documentclass{article}

\usepackage{microtype}
\usepackage{graphicx}
\usepackage{subfigure}
\usepackage{booktabs} 
\usepackage{amsmath}
\usepackage{amsthm}
\usepackage{amsfonts} 
\usepackage{mathtools}
\usepackage{graphbox} 
\usepackage{thmtools,thm-restate}

\usepackage{url}

\usepackage{breakurl}
\usepackage[breaklinks]{hyperref}

\usepackage{hyperref}

\usepackage[preprint]{icml2019}

\icmltitlerunning{Wasserstein Adversarial Examples via Projected Sinkhorn Iterations}

\DeclareMathOperator*{\maximize}{maximize}
\DeclareMathOperator*{\minimize}{minimize}

\DeclareMathOperator*{\subjectto}{subject\;to}

\DeclareMathOperator*{\argmin}{arg\;min}
\DeclareMathOperator*{\argmax}{arg\;max}
\DeclareMathOperator*{\proj}{proj}

\begin{document}

\twocolumn[
\icmltitle{Wasserstein Adversarial Examples via Projected Sinkhorn Iterations}




\begin{icmlauthorlist}
\icmlauthor{Eric Wong}{mld}
\icmlauthor{Frank R. Schmidt}{bosch}
\icmlauthor{J. Zico Kolter}{csd,bosch-us}
\end{icmlauthorlist}

\icmlaffiliation{mld}{Machine Learning Department, Carnegie Mellon University, Pittsburgh, Pennsylvania, USA}
\icmlaffiliation{bosch}{Bosch Center for Artificial Intelligence, Renningen, Germany}
\icmlaffiliation{bosch-us}{Bosch Center for Artificial Intelligence, Pittsburgh, Pennsylvania, USA}
\icmlaffiliation{csd}{Computer Science Department, Carnegie Mellon University, Pittsburgh, Pennsylvania, USA}

\icmlcorrespondingauthor{Eric Wong}{ericwong@cs.cmu.edu}

\icmlkeywords{Machine Learning, ICML}

\vskip 0.3in
]



\printAffiliationsAndNotice{}  

\begin{abstract}
A rapidly growing area of work has studied the existence of adversarial examples, 
datapoints which have been perturbed to fool a classifier, but the vast 
majority of these works have focused primarily on threat models 
defined by $\ell_p$ norm-bounded perturbations. In this paper, we propose a new
threat model for adversarial attacks based on the Wasserstein distance.  In the image classification setting, such distances measure the cost of moving pixel mass, which naturally cover ``standard'' image manipulations such as scaling, rotation, translation, and distortion (and can potentially be applied to other settings as well).  To generate Wasserstein 
adversarial examples, we develop a procedure for projecting onto the 
Wasserstein ball, based upon a modified version of the Sinkhorn iteration.  The resulting algorithm 
can successfully attack image classification models, 
bringing traditional CIFAR10 models down to 
3\% accuracy within a Wasserstein ball with radius 0.1 (i.e., moving 10\% of the image mass 1 pixel), 
and we demonstrate that PGD-based adversarial training 
can improve this adversarial accuracy to 76\%.
In total, this work opens up a new direction of study in adversarial robustness, more formally considering convex metrics that accurately capture the invariances that we typically believe should exist in classifiers. Code for all experiments in the paper is available at \url{https://github.com/locuslab/projected_sinkhorn}. 
\end{abstract}

\section{Introduction}
A substantial effort in machine learning research has gone towards studying 
\emph{adversarial examples} \citep{szegedy2014intriguing}, 
commonly described as datapoints that 
are indistinguishable from ``normal'' examples, but are specifically perturbed 
to be misclassified by machine learning systems. This notion of 
indistinguishability, later described as the threat model for attackers, 
was originally taken to be $\ell_\infty$ bounded perturbations, which model a 
small amount of noise injected to each pixel \citep{goodfellow2015explaining}.  
Since then, subsequent work on understanding, attacking, and defending against 
adversarial examples has largely focused on this $\ell_\infty$ threat model 
and its corresponding $\ell_p$ generalization. 
While the $\ell_p$ ball is a convenient source of 
adversarial perturbations, it is by no means a comprehensive 
description of all possible adversarial perturbations. Other work 
\citep{engstrom2017rotation} has looked at perturbations such as rotations and translations, but beyond these specific transforms, there has been little work considering broad classes of attacks beyond the $\ell_p$ ball.

\begin{figure}[t]
\vskip -0.1in
\begin{center}
\begin{equation*}
\begin{alignedat}{2}
\begin{gathered}
\includegraphics[width=0.4\columnwidth]{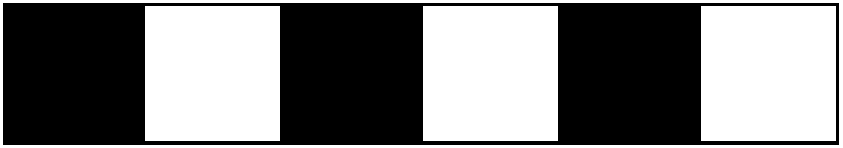}
\end{gathered} & + \Delta_W &=
\begin{gathered}
\includegraphics[width=0.4\columnwidth]{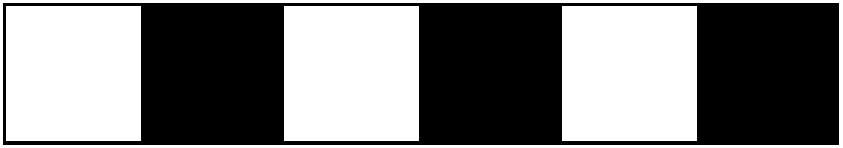}
\end{gathered}\\
\begin{gathered}
\includegraphics[width=0.4\columnwidth]{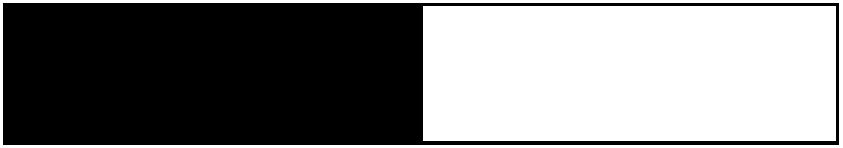}
\end{gathered} & + \Delta_\infty &=
\begin{gathered}
\includegraphics[width=0.4\columnwidth]{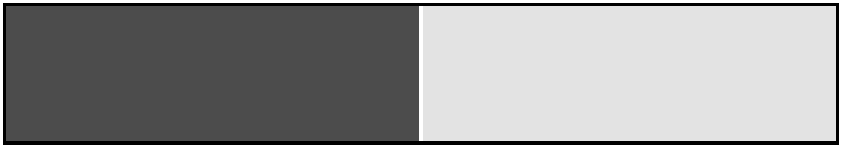}
\end{gathered}
\end{alignedat}
\end{equation*}
\vskip -0.2in
\caption{A minimal example exemplifying the difference between Wasserstein perturbations and 
$\ell_\infty$ perturbations on an image with six pixels. The top example utilizes a perturbation 
$\Delta_W$ to shift the image one pixel to the right, which is small 
with respect to Wasserstein distance since each pixel moved a minimal amount, 
but large with respect to $\ell_\infty$ distance since each pixel changed a maximal amount. 
In contrast, the bottom example utilizes a perturbation $\Delta_\infty$ which changes all 
pixels to be grayer. This is small with respect to $\ell_\infty$ distance, since each pixel changes 
by a small amount, but large with respect to Wasserstein distance, since the mass on each pixel 
on the left had to move halfway across the image to the right.
}
\label{fig:minimalexample}
\end{center}
\vskip -0.2in
\end{figure}


In this paper, we propose a new type of adversarial perturbation that encodes a 
general class of attacks that is fundamentally different 
from the $\ell_p$ ball. 
Specifically, we propose an attack model where the perturbed examples are 
bounded in Wasserstein distance from the original example. This distance
can be intuitively understood for images as the cost of moving around pixel mass 
to move from one image to another. 
Note that the Wasserstein ball and the $\ell_p$ ball can be quite different 
in their allowable perturbations: examples that are close in Wasserstein distance 
can be quite far in $\ell_p$ distance, and vice versa (a pedagogical example 
demonstrating this is in Figure \ref{fig:minimalexample}). 

We develop this idea 
of Wasserstein adversarial examples in two main ways. 
Since adversarial examples are typically best generated using variants of projected gradient descent, we first derive an algorithm that projects onto the Wasserstein ball. However, performing an exact projection is computationally expensive, so our main contribution here is to derive a fast method for \emph{approximate} projection. The procedure can be viewed as a modified Sinkhorn iteration, but with a more complex set of update equations.   Second, we develop efficient methods for adversarial training under this threat method.  Because this involves repeatedly running this projection within an inner optimization loop, speedups that use a \emph{local} transport plan are particularly crucial (i.e. only moving pixel mass to nearby pixels), making the projection complexity linear in the image size.

We evaluate the attack quality on standard models, showing for example that we can reduce the adversarial accuracy of a standard CIFAR10 classifier from 94.7\% to 3\% using a Wasserstein ball of radius 0.1 (equivalent to moving 10\% of the mass of the image by one pixel), whereas the same attack reduces the adversarial accuracy of a model certifiably trained against $\ell_\infty$ perturbations from 66\% to 61\%.  In contrast, we show that with adversarial training, we are able to improve the adversarial accuracy of this classifier to 76\% while retaining a nominal accuracy of 80.7\%.  We additionally show, however, that existing \emph{certified} defenses cannot be easily extended to this setting; building models provably robust to Wasserstein attacks will require fundamentally new techniques.  In total, we believe this work highlights a new direction in adversarial examples: convex perturbation regions which capture a much more intuitive form of structure in their threat model, and which move towards a more ``natural'' notion of adversarial attacks.

\section{Background and Related Work}
Much of the work in adversarial examples has focused on the original $\ell_\infty$ 
threat model presented by \citet{goodfellow2015explaining}, some of which 
also extends naturally to $\ell_p$ perturbations. 
Since then, there has been a plethora of 
papers studying this threat model, ranging from improved attacks, 
heuristic and certified defenses, and verifiers. As there are far too many 
to discuss here, we highlight a few which are the most relevant to this work. 

The most commonly used method for generating adversarial examples 
is to use a form of projected gradient descent over 
the region of allowable perturbations, originally referred to as 
the Basic Iterative Method \citep{kurakin2017adversarial}. Since then, 
there has been a back-and-forth of new heuristic defenses followed by 
more sophisticated attacks. To name a few, distillation was 
proposed as a defense but was defeated \citep{papernot2016distillation, carlini2017towards}, 
realistic transformations seen by vehicles were thought to be safe until 
more robust adversarial examples were created \cite{lu2017no, pmlr-v80-athalye18b}, 
and many defenses submitted to ICLR 2018 were broken before the review 
period even finished \citep{obfuscated-gradients}. 
One undefeated heuristic defense is to use the 
adversarial examples in adversarial training, 
which has so far worked well in practice \citep{madry2018towards}. 
While this method has traditionally been used for $\ell_\infty$ and $\ell_2$ balls 
(and has a natural $\ell_p$ generalization), 
in principle, the method can be used to project onto any kind of perturbation region. 


Another set of related papers are verifiers and provable defenses, which aim 
to produce (or train on) certificates that are provable guarantees of robustness 
against adversarial attacks. Verification methods are now applicable to multi-layer 
neural networks using techniques ranging from semi-definite 
programming relaxations \cite{raghunathan2018semi}, mixed integer linear programming \cite{tjeng2018evaluating}, 
and duality \cite{dvijotham18}. 
Provable defenses are able to tie verification into training non-trivial deep networks 
by backpropagating through certificates, which are generated with 
duality-based bounds \citep{wong2018provable, wong2018scaling}, 
abstract interpretations \citep{mirman2018diff}, and interval bound propagation \cite{gowal2018interval}. 
These methods have subsequently inspired new heuristic training defenses, where the resulting models 
can be independently verified as robust \citep{croce2018provable, xiao2018training}. 
Notably, some of these approaches are \emph{not} overly reliant on specific types of 
perturbations (e.g. duality-based bounds). Despite their generality, these certificates 
have only been trained and evaluated in the context of $\ell_\infty$ and $\ell_2$ balls, and 
we believe this is due in large part to a lack of alternatives. 

Highly relevant to this work are attacks that lie outside the traditional $\ell_p$ ball 
of imperceptible noise. 
For example, 
simple rotations and translations form a fairly limited set of 
perturbations that can be quite large in $\ell_p$ norm, but are sometimes sufficient 
in order to fool classifiers \citep{engstrom2017rotation}. 
On the other hand, adversarial examples 
that work in the real world do not necessarily conform 
to the notion of being ``imperceptible'', and need to utilize a 
stronger adversary that is visible to real world systems. Some examples include    
wearing adversarial 3D printed glasses to fool facial recognition 
\citep{sharif2017adversarial}, the use of 
adversarial graffiti to attack traffic sign classification \citep{eykholt2018robust}, 
and printing adversarial 
textures on objects to attack image classifiers \citep{pmlr-v80-athalye18b}. 
While \citet{sharif2017adversarial} allows perturbations that are physical glasses,
the others use an $\ell_p$ threat model with a larger radius, when a different 
threat model could be a more natural description of adversarial 
examples that are perceptible on camera. 

Last but not least, our paper relies heavily on the Wasserstein 
distance, which has seen 
applications throughout machine learning. The 
traditional notion of Wasserstein distance has the drawback of being 
computationally expensive: computing a single distance involves solving 
an optimal transport problem (a linear program) with a 
number of variables quadratic in the dimension of 
the inputs. However, it was shown that by subtracting an entropy regularization 
term, one can compute approximate Wasserstein distances extremely quickly 
using the Sinkhorn iteration
\citep{cuturi2013sinkhorn}, which was later shown to run in 
near-linear time \citep{altschuler2017near}. 
Relevant but orthogonal to our work, is that of \citet{sinha2018certifying} 
on achieving distributional robustness using the Wasserstein distance. While we both 
use the Wasserstein distance in the context of adversarial training, the 
approach is quite different: \citet{sinha2018certifying} use the Wasserstein distance 
to perturb the underlying \emph{data distribution}, 
whereas we use the Wasserstein distance as an attack model for perturbing 
each \emph{example}. 

\begin{figure}[t]
\begin{center}
\setlength{\tabcolsep}{2pt}
\begin{tabular}{ccccc}
\includegraphics[align=c,width=0.18\columnwidth]{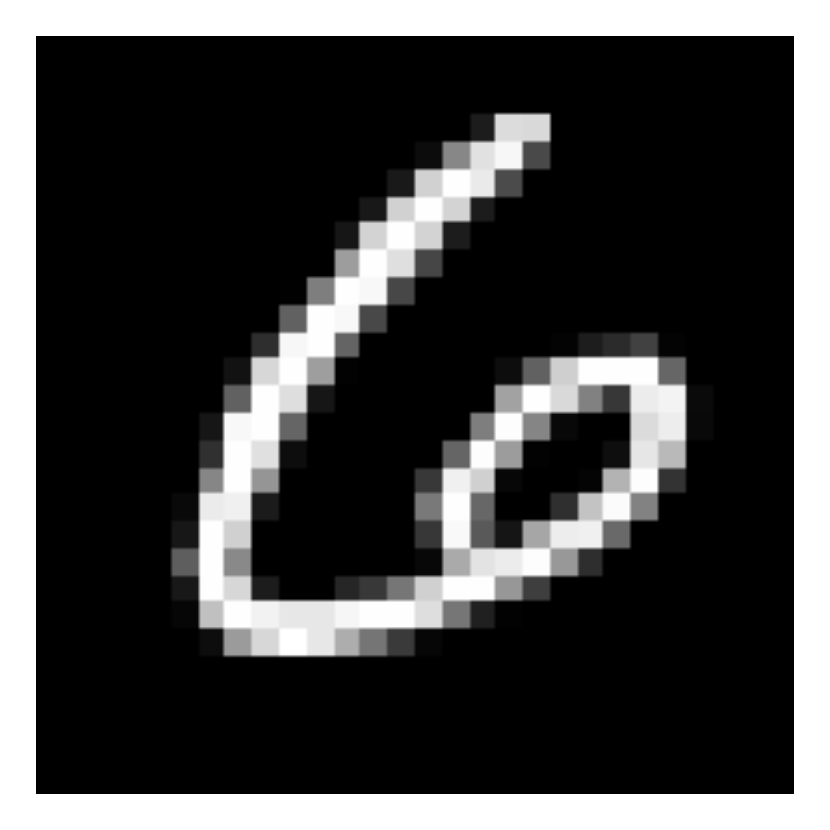} & $+$ & 
\includegraphics[align=c,width=0.18\columnwidth]{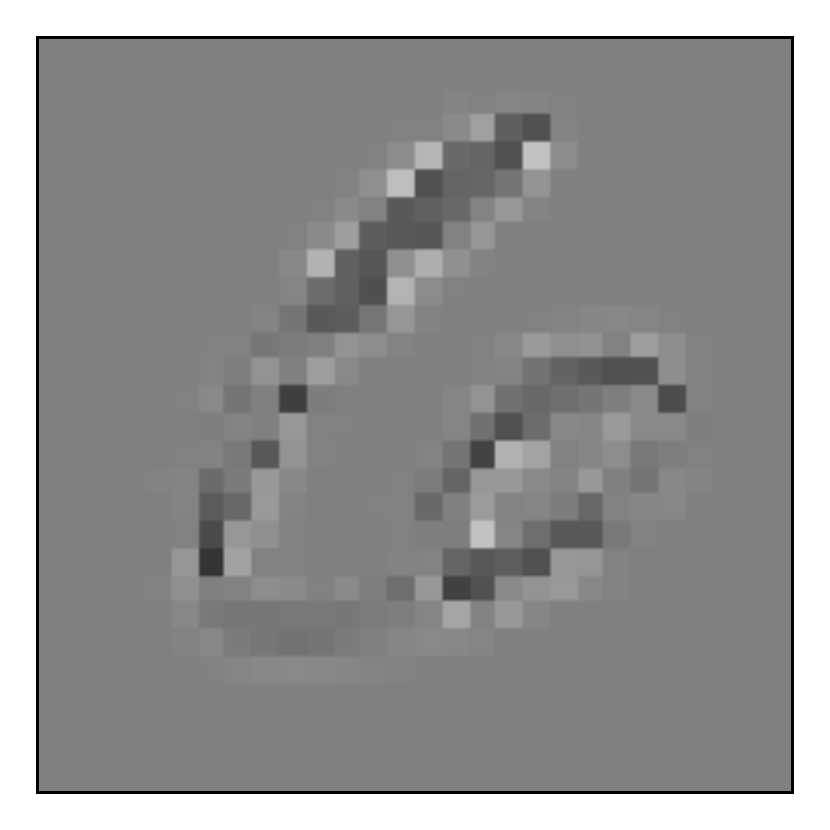} & $=$ & 
\includegraphics[align=c,width=0.18\columnwidth]{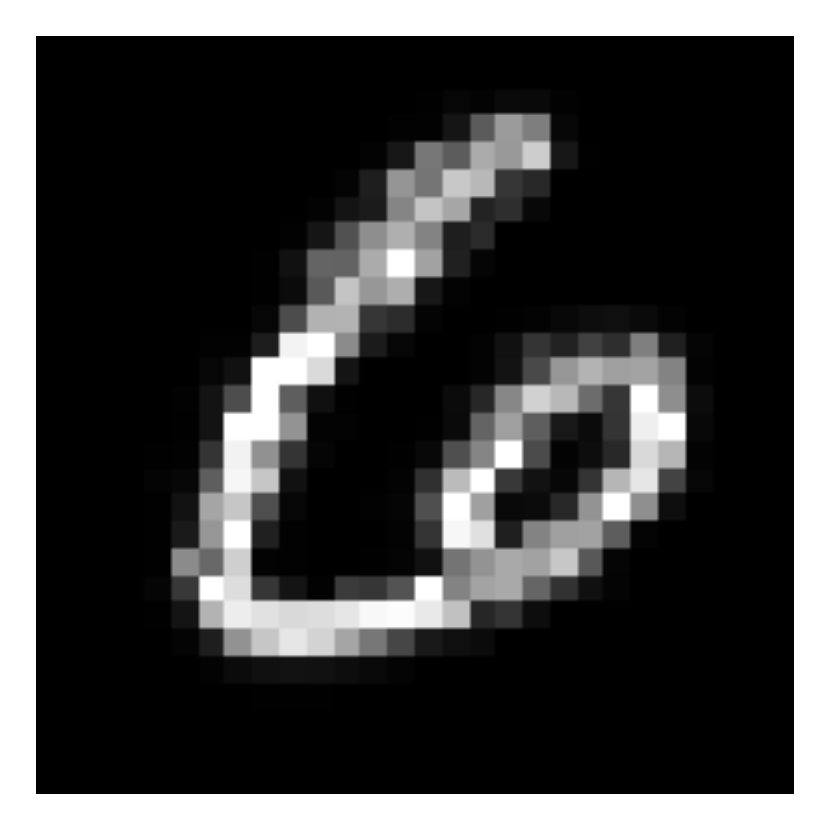}\\
\includegraphics[align=c,width=0.18\columnwidth]{figures/vs/six.pdf} & $+$ & 
\includegraphics[align=c,width=0.18\columnwidth]{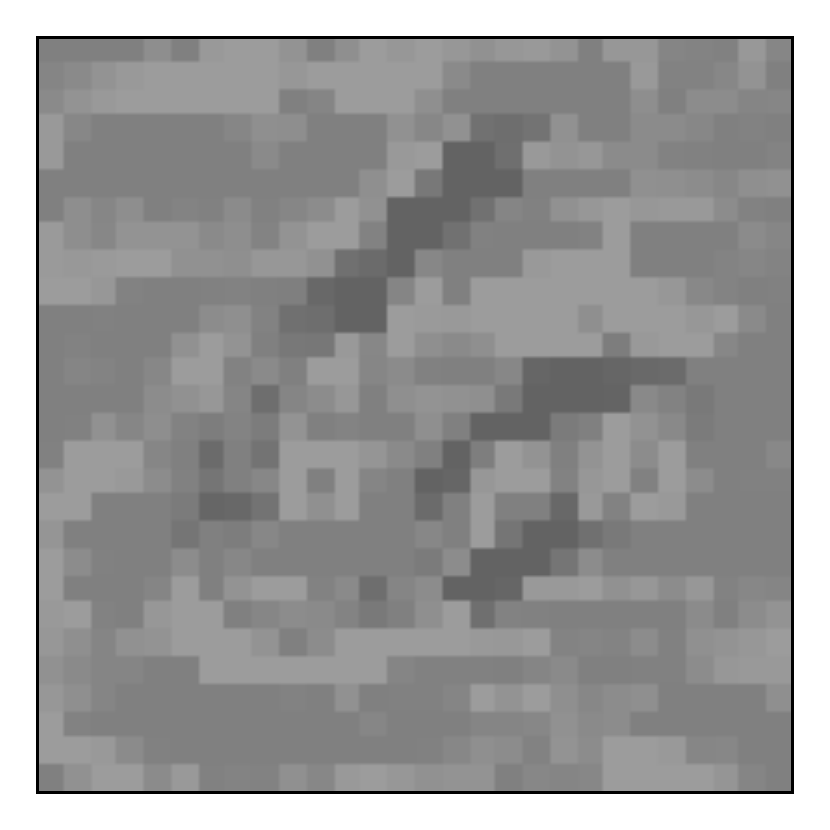} & $=$ & 
\includegraphics[align=c,width=0.18\columnwidth]{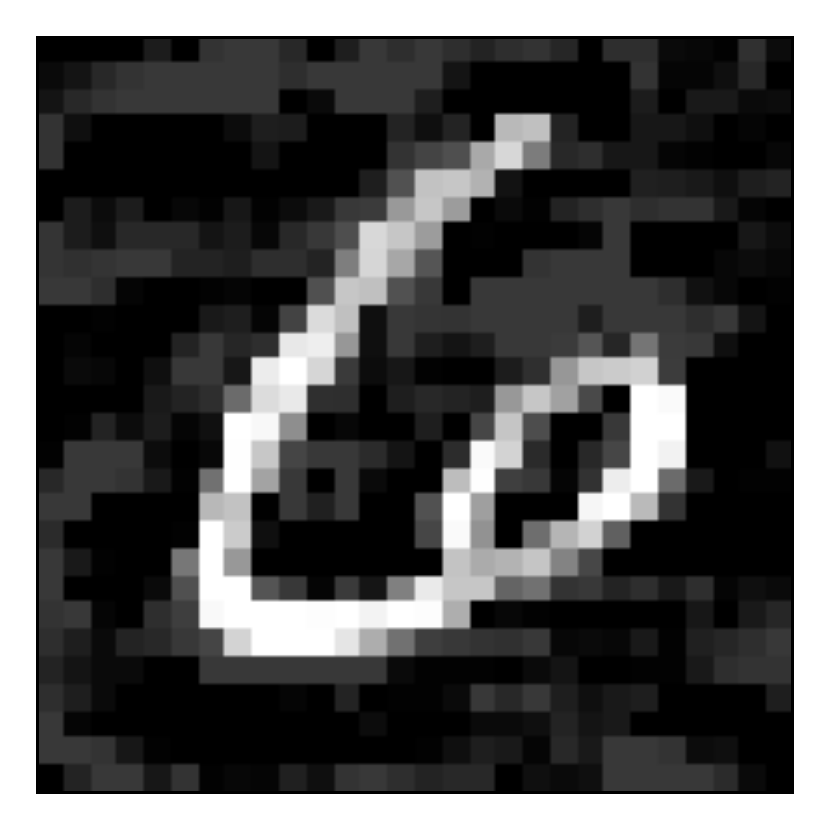}\\
\includegraphics[align=c,width=0.22\columnwidth]{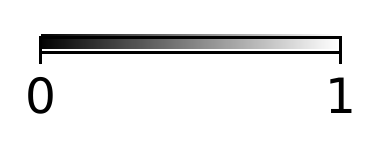} &  & 
\includegraphics[align=c,width=0.22\columnwidth]{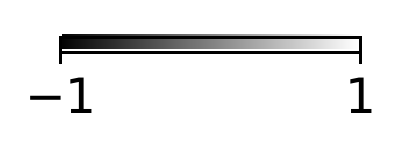}\hspace{2px} &  & 
\includegraphics[align=c,width=0.22\columnwidth]{figures/vs/colorbar_0_1.pdf}\\
\end{tabular}
\vskip -0.1in
\caption{A comparison of a Wasserstein (top) vs an $\ell_\infty$ (bottom) adversarial example 
for an MNIST classifier (for $\epsilon=0.4$ and $0.3$ respectively), 
showing the original image (left), the added perturbation (middle), and the 
final perturbed image (right). We find that the Wasserstein perturbation has a structure reflecting 
the actual content of the image, whereas the $\ell_\infty$ perturbation also attacks the 
background pixels. 
}
\label{fig:wasserstein_vs_infinity}
\end{center}
\vskip -0.2in
\end{figure}

\paragraph{Contributions} 
This paper takes a step back from using $\ell_p$ as a perturbation metric, and 
proposes using the Wasserstein distance instead as an equivalently general 
but qualitatively different way of generating adversarial examples. To tackle 
the computational complexity of projecting onto a Wasserstein ball, 
we use ideas from the Sinkhorn iteration \citep{cuturi2013sinkhorn} 
to derive a fast method for an approximate projection. Specifically, we show 
that subtracting a similar entropy-regularization term to the projection problem 
results in a Sinkhorn-like algorithm, and using local transport plans makes 
the procedure tractable for generating adversarial images. In contrast to 
$\ell_\infty$ and $\ell_2$ perturbations, we find that the Wasserstein metric 
generates adversarial examples whose perturbations have inherent structure reflecting 
the actual image itself (see Figure \ref{fig:wasserstein_vs_infinity} for a comparison). 
We demonstrate the efficacy of this attack on 
standard models, models trained against this attack, and provably robust models 
(against $\ell_\infty$ attacks) on MNIST and CIFAR10 datasets. 
While the last of these models are not trained to be 
robust specifically against this attack, we observe that 
that some (but not all) robustness empirically transfers over 
to protection against the Wasserstein attack. 
More importantly, we show that while the Wasserstein ball does fit naturally into 
duality based frameworks for generating and training against certificates, 
there is a fundamental roadblock preventing  
these methods from generating non-vacuous bounds on Wasserstein balls. 

\section{Preliminaries}

\paragraph{PGD-based adversarial attacks} 
The most common method of creating adversarial examples is to use 
a variation of projected gradient descent. 
Specifically, let $(x,y)$ be a datapoint and its label, and 
let $\mathcal B(x,\epsilon)$ 
be some ball around $x$ with radius $\epsilon$, which represents the 
threat model for the adversary. 
We first define the projection operator 
onto $\mathcal{B}(x,\epsilon)$ to be 
\begin{equation}
\proj_{\mathcal B(x,\epsilon)}(w) = \argmin_{z \in B(x,\epsilon)} \|w - z\|_2^2
\end{equation}
which finds the point closest (in Euclidean space) to the input $w$ that lies within 
the ball $\mathcal B(x,\epsilon)$. Then, 
for some step size $\alpha$ and some loss $\ell$ (e.g. cross-entropy loss), 
the algorithm consists of the following iteration: 
\begin{equation}
x^{(t+1)} = \proj_{\mathcal B(x, \epsilon)}\left(x^{(t)} + \argmax_{\|v\|\leq \alpha}v^T\nabla \ell(x^{(t)},y)\right)
\label{eq:pgd}
\end{equation}
where $x^{(0)} = x$ or any randomly initialized point within $\mathcal B(x,\epsilon)$. 
This is sometimes referred to as projected \emph{steepest} descent, which is used to generated 
adversarial examples since the standard gradient steps are typically too small. 
If we consider the $\ell_\infty$ ball $\mathcal{B}_\infty(x,\epsilon) = \{x + \Delta : \|\Delta\|_\infty \leq \epsilon\}$ and use steepest descent with respect to the $\ell_\infty$ norm, then we 
recover the Basic Iterative Method originally presented by \citet{kurakin2017adversarial}. 

\paragraph{Adversarial training}
One of the heuristic defenses that works well in practice is to use adversarial 
training with a PGD adversary. 
Specifically, instead of minimizing the loss 
evaluated at a example $x$, we minimize the loss on an adversarially 
perturbed example $x_{adv}$, where $x_{adv}$ is obtained by running the projected 
gradient descent attack for the ball $\mathcal B(x,\epsilon)$ for some number of iterations, as shown in Algorithm \ref{alg:adv_training}. Taking $\mathcal B(x,\epsilon)$ to be an $\ell_\infty$  
ball recovers the procedure used by \citet{madry2018towards}. 

\begin{algorithm}[tb]
   \caption{An epoch of adversarial training for a loss function $\ell$, classifier $f_\theta$ with parameters $\theta$, and step size parameter $\alpha$ for some ball $\mathcal B$.}
\label{alg:adv_training}
\begin{algorithmic}
   \STATE {\bfseries input:} Training data $(x_i,y_i)$, $i=1\dots n$
   \FOR{$i=1\dots n$}
   \STATE \textit{// Run PGD adversary}
   \STATE $x_{adv} \coloneqq x_i$
   \FOR{$t=1\dots T$}
   \STATE $\delta \coloneqq \argmax_{\|v\|\leq \alpha}v^T\nabla \ell(x_{adv},y_i)$
   \STATE $x_{adv} \coloneqq \proj_{\mathcal B(x_i, \epsilon)}\left(x_{adv} + \delta\right)$
   \ENDFOR
   \STATE \textit{// Backpropagate with $x_{adv}$, e.g. with SGD}
   \STATE Update $\theta$ with $\nabla \ell(f_\theta(x_{adv}), y_i)$
   \ENDFOR
\end{algorithmic}
\end{algorithm}

\paragraph{Wasserstein distance}
Finally, we define the most crucial component of this work, an alternative 
metric from $\ell_p$ distances. 
The Wasserstein distance (also referred to as the Earth 
mover's distance) is an optimal transport problem
that can be intuitively understood in the 
context of distributions as the minimum cost of 
moving probability mass to change one distribution 
into another. When applied to images, this can 
be interpreted as the cost of moving pixel mass from one pixel to 
another another, where the cost increases with distance.

More 
specifically, let $x,y\in \mathbb R^n_+$ be two non-negative data points such 
that $\sum_i x_i = \sum_j y_j = 1$, so images and other inputs need to be normalized, 
and let $C\in \mathbb R^{n\times n}_+$ be some non-negative 
cost matrix where $C_{ij}$ 
encodes the cost of moving mass from $x_i$ to $y_j$. 
Then, the Wasserstein distance $d_\mathcal W$ 
between $x$ and $y$ is defined to be 
\begin{equation}
\begin{split}
d_\mathcal{W}(x,y) =& \min_{\Pi\in \mathbb{R}^{n\times n}_+} \langle \Pi, C\rangle\\
\subjectto & \;\;\Pi1 = x, \;\;\Pi^T1 = y
\end{split}
\label{eq:wasserstein}
\end{equation}
where the minimization over transport plans $\Pi$, whose entries 
$\Pi_{ij}$ encode how the mass moves from $x_i$ to $y_j$. Then, we can 
define the Wasserstein ball with radius $\epsilon$ as 
\begin{equation}
\mathcal B_\mathcal{W}(x,\epsilon) = \{x + \Delta : d_\mathcal W(x,x+\Delta) \leq \epsilon\}
\label{eq:wasserstein_ball}
\end{equation}

\section{Wasserstein Adversarial Examples}
The crux of this work relies on offering a fundamentally different 
type of adversarial example from typical, 
$\ell_p$ perturbations: the Wasserstein adversarial example.




\subsection{Projection onto the Wasserstein Ball}
In order to generate Wasserstein adversarial examples, we can run the projected 
gradient descent attack from Equation \eqref{eq:pgd}, dropping in the Wasserstein ball 
$\mathcal{B}_\mathcal{W}$ from Equation \eqref{eq:wasserstein_ball} in place of $\mathcal B$. 
However, while 
projections onto regions such as $\ell_\infty$ and $\ell_2$ balls 
are straightforward and have closed form computations, 
simply computing the Wasserstein distance itself requires solving 
an optimization problem. Thus, 
the first natural requirement to generating Wasserstein adversarial 
examples is to derive an \emph{efficient} way to project 
examples onto a Wasserstein ball of radius 
$\epsilon$. Specifically, projecting $w$ onto the Wasserstein ball 
around $x$ with radius $\epsilon$ and transport cost matrix $C$ 
can be written as solving the following optimization problem: 
\begin{equation}
\begin{split}
\minimize_{z\in\mathbb R^n_+, \Pi\in \mathbb{R}^{n\times n}_+} &\;\; \frac{1}{2}\|w-z\|_2^2\\
\subjectto \;\; & \Pi1 = x,\;\; \Pi^T1 = z\\
& \langle \Pi,C\rangle \leq \epsilon 
\end{split}
\label{eq:projection}
\end{equation}
While we could directly solve this optimization problem (using an 
off-the-shelf quadratic programming solver), this is prohibitively expensive 
to do for every iteration of projected gradient descent, especially since 
there is a quadratic number of variables. 
However, \citet{cuturi2013sinkhorn} showed that the 
standard Wasserstein distance problem from Equation \eqref{eq:wasserstein} 
can be approximately solved efficiently 
by subtracting an entropy regularization term on the transport plan $W$, and 
using the Sinkhorn-Knopp matrix scaling algorithm. 
Motivated by these results, instead of solving the projection 
problem in Equation \eqref{eq:projection} exactly, the key contribution 
that allows us to do the projection efficiently is to instead 
solve the following entropy-regularized projection problem: 
 \begin{equation}
\begin{split}
\minimize_{z\in\mathbb R^n_+, \Pi\in \mathbb{R}^{n\times n}_+} &\;\; \frac{1}{2}\|w-z\|_2^2 + \frac{1}{\lambda}\sum_{ij}\Pi_{ij}\log(\Pi_{ij})\\
\subjectto \;\; & \Pi1 = x,\;\; \Pi^T1 = z\\
& \langle \Pi,C\rangle \leq \epsilon. 
\end{split}
\label{eq:projection_reg}
\end{equation}
Although this is an \emph{approximate} projection onto the Wasserstein ball, importantly, 
the looseness in the approximation is only in finding the projection $z$ which is closest 
(in $\ell_2$ norm) to the original example $x$. All feasible points, including the 
optimal solution, are still within the actual $\epsilon$-Wasserstein ball, so examples generated 
using the approximate projection are still within the Wasserstein threat model. 

Using the method of Lagrange multipliers, we can introduce dual variables $(\alpha, \beta, \psi)$ 
and derive an equivalent dual problem in 
Lemma \ref{lem:dual} (the proof is deferred to Appendix \ref{app:dual}). 
\begin{restatable}{lemma}{duallemma}
The dual of the entropy-regularized Wasserstein projection problem in Equation \eqref{eq:projection_reg} 
is
\begin{equation}
\maximize_{\alpha, \beta \in \mathbb R^n, \psi \in \mathbb R_+} g(\alpha, \beta, \psi)
\end{equation} 
where
\begin{equation}
\begin{split}
g(\alpha, \beta,\psi) =& -\frac{1}{2\lambda}\|\beta\|_2^2 - \psi \epsilon + \alpha^Tx + \beta^Tw \\
 - \sum_{ij}&\exp(\alpha_i)\exp( - \psi C_{ij} - 1)\exp(\beta_j)\\
\end{split}
\label{eq:dual}
\end{equation}
\label{lem:dual}
\end{restatable}
Note that the dual problem here differs from the traditional dual problem for Sinkhorn iterates by 
having an additional quadratic term on $\beta$ and an additional dual variable $\psi$. 
Nonetheless, we can still derive a Sinkhorn-like algorithm by performing block coordinate 
ascent over the dual variables (the full derivation can be found in Appendix \ref{app:dualexplain}). Specifically, maximizing $g$ with respect to 
$\alpha$ results in 
\begin{equation}
\begin{gathered}
\quad \argmax_{\alpha_i}g(\alpha,\beta,\psi)=\\
\log \left(x_i\right) - \log \left(\sum_{j}\exp( - \psi C_{ij} - 1)\exp( \beta_j)\right),
\label{eq:sinkhorn_iterate}
\end{gathered}
\end{equation}
which is identical (up to a log transformation of variables) to the original Sinkhorn 
iterate proposed in \citet{cuturi2013sinkhorn}. The maximization step for $\beta$ can also 
be done analytically with 
\begin{equation}
\begin{gathered}
\quad \argmax_{\beta_j}g(\alpha,\beta,\psi) =\\
\lambda w_j - W\left(\lambda \exp(\lambda w_j)\sum_{i}\exp(\alpha_i)\exp( - \psi C_{ij} - 1)\right)
\end{gathered}
\end{equation}
where $W$ is the Lambert $W$ function, which is defined as the inverse of $f(x) = x e^x$. 
Finally, since $\psi$ cannot be solved for analytically, we can perform the following Newton step 
\begin{equation}
\psi' = \psi - t\cdot \frac{\partial g / \partial \psi}{\partial^2 g / \partial \psi^2}
\end{equation}
where 
\begin{equation}
\begin{split}
\partial g / \partial \psi &= -\epsilon + \sum_{ij} \exp(\alpha_i)C_{ij}\exp( - \psi C_{ij})\exp(\beta_j)\\
\partial^2 g / \partial \psi^2 &= -\sum_{ij} \exp(\alpha_i)C_{ij}^2\exp( - \psi C_{ij})\exp(\beta_j)
\end{split}
\end{equation}
and where $t$ is small enough such that $\psi' \geq 0$. 
Once we have solved the dual problem, we can recover the primal solution 
(to get the actual projection), which is 
described in Lemma \ref{lem:dualtoprimal} and proved in Appendix \ref{app:dualtoprimal}. 
\begin{restatable}{lemma}{dualtoprimal}
Suppose $\alpha^*, \beta^*,\psi^*$ maximize the dual problem $g$ in Equation \eqref{eq:dual}. Then, 
\begin{equation}
\begin{split}
z_i^* &= w_i - \beta_i/\lambda\\
\Pi_{ij}^* &= \exp(\alpha^*_i)\exp(-\psi^* C_{ij} - 1)\exp(\beta^*_j)
\end{split}
\end{equation}
are the corresponding solutions that minimize the primal problem in Equation \eqref{eq:projection_reg}. 
\label{lem:dualtoprimal}
\end{restatable}
The whole algorithm can then be vectorized and implemented as  
Algorithm \ref{alg:projected_sinkhorn}, which we call 
projected Sinkhorn iterates. The algorithm uses a simple line search to ensure 
that the constraint $\psi \geq 0$ is not violated. Each iteration
has 8 $O(n^2)$ operations (matrix-vector product or matrix-matrix element-wise product), 
in comparison to the original Sinkhorn iteration which has 2 matrix-vector products.
\begin{algorithm}[tb]
   \caption{Projected Sinkhorn iteration to project $x$ onto the $\epsilon$ Wasserstein ball 
   around $y$. We use $\cdot$ to denote element-wise multiplication. The $\log$ and $\exp$ operators also apply element-wise. }
\label{alg:projected_sinkhorn}
\begin{algorithmic}
   \STATE {\bfseries input:} $x,w\in \mathbb R^n, C\in \mathbb C^{n\times n}, \lambda \in \mathbb R$
   \STATE Initialize $\alpha_i, \beta_i \coloneqq \log(1/n)$ for $i=1,\dots,n$ and $\psi\coloneqq 1$
   \STATE $u,v \coloneqq \exp(\alpha), \exp(\beta)$
   \WHILE{$\alpha,\beta,\psi$ not converged}
   \STATE \textit{// update $K$}
   \STATE $K_\psi \coloneqq \exp(-\psi C - 1)$
   \STATE 
   \STATE \textit{// block coordinate descent iterates}
   \STATE $\alpha \coloneqq \log(x) - \log(K_\psi v)$ 
   \STATE $u \coloneqq \exp(\alpha)$
   \STATE $\beta \coloneqq \lambda w - W\left(u^T K_\psi\cdot \lambda \exp(\lambda w) \right)$
   \STATE $v \coloneqq \exp(\beta)$
   \STATE 
   \STATE \textit{// Newton step}
   \STATE $g \coloneqq -\epsilon + u^T(C\cdot K_\psi)v$
   \STATE $h \coloneqq - u^T(C \cdot C \cdot K_\psi)v$
   \STATE
   \STATE \textit{// ensure }$\psi \geq 0$
   \STATE $\alpha \coloneqq 1$
   \WHILE{$\psi - \alpha g/h < 0$}
   \STATE $\alpha \coloneqq \alpha/2$
   \ENDWHILE
   \STATE $\psi \coloneqq \psi -\alpha g/h$
   \ENDWHILE
   \STATE {\bfseries return:} $w - \beta/\lambda$
\end{algorithmic}
\end{algorithm}

\paragraph{Matrix scaling interpretation}
The original Sinkhorn iteration has a natural interpretation as a matrix scaling 
algorithm, iteratively rescaling the rows and columns of a matrix to achieve 
the target distributions. The Projected Sinkhorn iteration has a similar interpretation: 
while the $\alpha$ step 
rescales the rows of $\exp(-\psi C - 1)$ to sum  to $x$, the $\beta$ 
step rescales the columns of $\exp(-\psi C - 1)$ to sum to $-\beta/\lambda + w$, which 
is the primal transformation of the projected variable $z$ at 
optimality as described in Lemma \ref{lem:dualtoprimal}. 
Lastly, the $\psi$ step can be interpreted as correcting for the transport cost of the current scaling: 
the numerator of the Newton step is simply the difference between the 
transport cost of the current matrix scaling and the maximum constraint $\epsilon$. A full 
derivation of the algorithm and a more detailed explanation 
on this interpretation can be found in Appendix \ref{app:dualexplain}. 


\subsection{Local Transport Plans}
The quadratic runtime dependence on input dimension can grow quickly, and this 
is especially true for images. Rather than allowing transport plans to move mass 
to and from any pair of pixels, we instead restrict the transport plan to move 
mass only within a $k \times k$ region of the originating pixel, 
similar in spirit to a convolutional filter. As a result, the cost matrix $C$ 
only needs to define the cost within a $k\times k$ region, and we can utilize tools 
used for convolutional filters to efficiently apply the cost to each $k\times k$ 
region. This reduces the computational complexity of each iteration to $O(nk^2)$. 
For images with more than one channel, we can use the same transport plan for 
each channel and only allow transport within a channel, so the cost matrix remains 
$k\times k$. For $5 \times 5$ local transport plans on CIFAR10, the projected Sinkhorn iterates 
typically converge in around 30-40 iterations, taking about 0.02 seconds per iteration 
on a Titan X for minibatches of size 100. Note that if we use a cost matrix $C$ that 
reflects the 1-Wasserstein distance, then this problem could be solved even more 
efficiently using Kantrovich duality, however we use this formulation to enable more 
general $p$-Wasserstein distances, or even non-standard cost matrices. 

\paragraph{Projected gradient descent on the Wasserstein ball}
With local transport plans, the method is fast enough to be used within a 
projected gradient descent routine to generate adversarial examples on images, 
and further used for adversarial training as in Algorithm \ref{alg:adv_training} 
(using steepest descent with respect to $\ell_\infty$ norm), 
except that we do an approximate projection onto the Wasserstein ball using 
Algorithm \ref{alg:projected_sinkhorn}.

%



\section{Results}
In this section, we run the Wasserstein examples through a range of 
typical experiments in the literature of adversarial examples. 
Table \ref{tab:nominal} 
summarizes the nominal error rates obtained by all considered models. 
All experiments can be run on a single GPU, and 
all code for the experiments is available at \url{https://github.com/locuslab/projected_sinkhorn}. 

\begin{table}[t]
\caption{Classification accuracies for models used in the experiments. }
\label{tab:nominal}
\vskip 0.15in
\begin{center}
\begin{small}
\begin{sc}
\begin{tabular}{llccr}
\toprule
Data set & Model & Nominal Accuracy  \\
\midrule
MNIST     & Standard      & 98.90\% \\
          & Binarize      & 98.73\% \\
          & Robust        & 98.20\% \\
          & Adv. Training & 96.95\% \\
\midrule
CIFAR10   & Standard      & 94.70\%\\
          & Robust        & 66.33\%\\
          & Adv. Training & 80.69\%\\
\bottomrule
\end{tabular}
\end{sc}
\end{small}
\end{center}
\vskip -0.2in
\end{table}

\paragraph{Architectures} For MNIST we used the convolutional 
ReLU architecture used in \citet{wong2018provable}, with two convolutional layers 
with 16 and 32 $4\times 4$ filters each, followed by a fully connected layer with 100 
units, which achieves a nominal accuracy of 98.89\%. 
For CIFAR10 we focused on the standard ResNet18 architecture \citep{he2016deep}, which 
achieves a nominal accuracy of 94.76\%. 

\paragraph{Hyperparameters}
For all experiments in this section, we focused on using $5 \times 5$ local transport plans 
for the Wasserstein ball, and used an entropy regularization constant of 1000 for 
MNIST and 3000 for CIFAR10. The cost matrix used for transporting between pixels 
is taken to be the 2-norm of the distance in pixel space (e.g. the cost of going from 
pixel $(i,j)$ to $(k,l)$ is $\sqrt{|i-j|^2 + |k-l|^2}$), which makes the optimal transport cost 
a metric more formally known as the 1-Wasserstein distance. 
For more extensive experiments on using different sizes of transport plans, 
different regularization constants, and different cost matrices, we 
direct the reader to Appendix \ref{app:experiments}. 

\paragraph{Evaluation at test time}
We use the follow evaluation procedure to attack 
models with projected gradient descent on the Wasserstein ball. 
For each MNIST example, we start with $\epsilon=0.3$ and increase it by a factor of 1.1 
every 10 iterations until either an adversarial example is found or until 200 iterations 
have passed, allowing for a maximum perturbation radius of $\epsilon=2$. For CIFAR10, 
we start with $\epsilon=0.001$ and increase it by a factor of 1.17 until either 
and adversarial example is found or until 400 iterations have passed, allowing for a maximum 
perturbation radius of $\epsilon=0.53$. 

\subsection{MNIST}
\begin{figure}[t]
\begin{center}
\setlength{\tabcolsep}{0pt}
\begin{tabular}{ccccc}
\includegraphics[align=c,width=0.2\columnwidth]{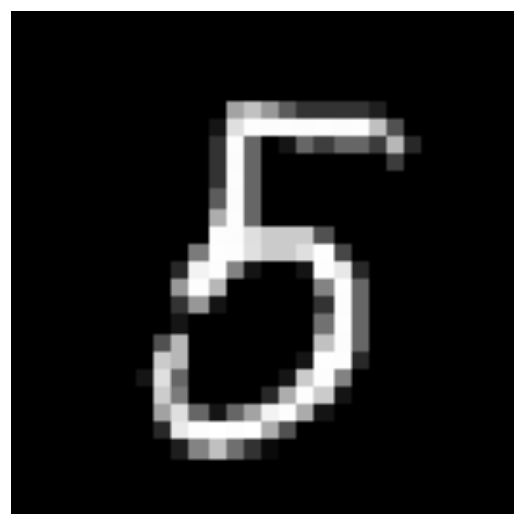} 
& \;+\; &
\begin{tabular}{c}
\includegraphics[align=c,width=0.12\columnwidth]{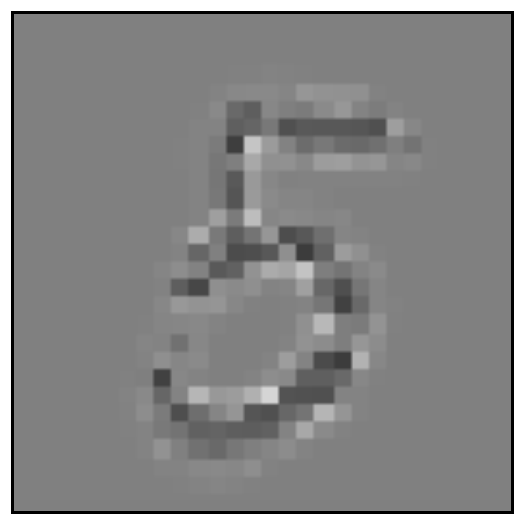} \\
\includegraphics[align=c,width=0.12\columnwidth]{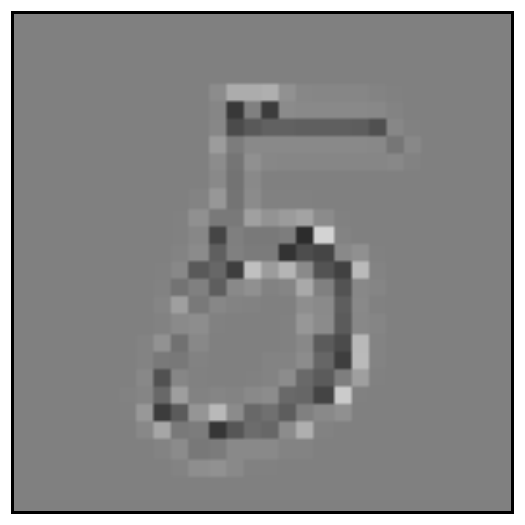} \\
\includegraphics[align=c,width=0.12\columnwidth]{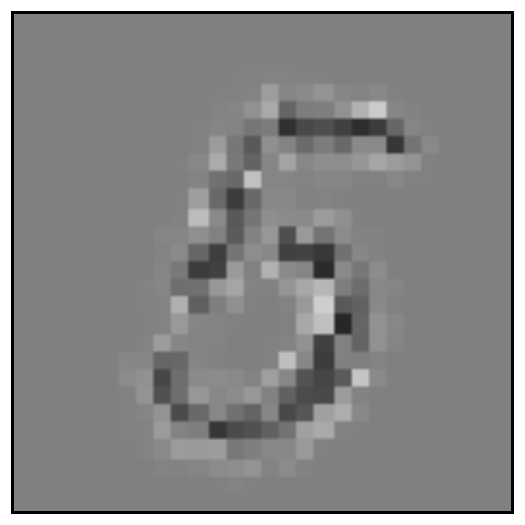} \\
\includegraphics[align=c,width=0.12\columnwidth]{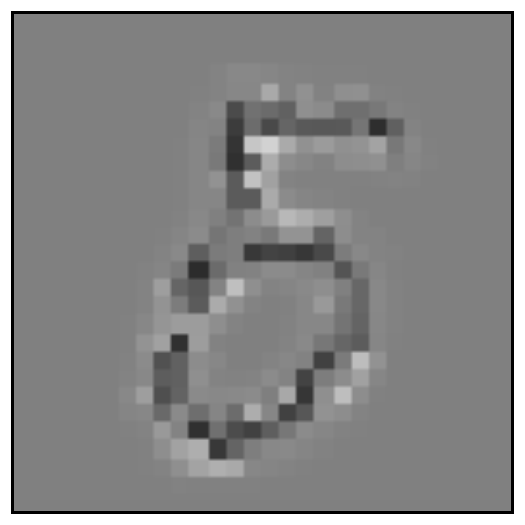} \\
\end{tabular}
& \;=\; & 
\begin{tabular}{l}
\includegraphics[align=c,width=0.12\columnwidth]{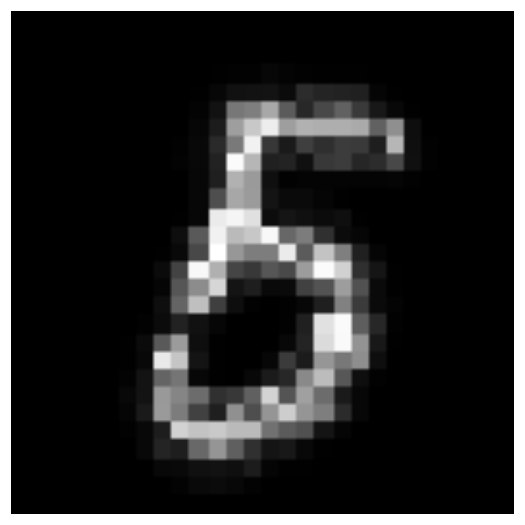} \quad {\footnotesize standard, $\epsilon=0.53$}\\
\includegraphics[align=c,width=0.12\columnwidth]{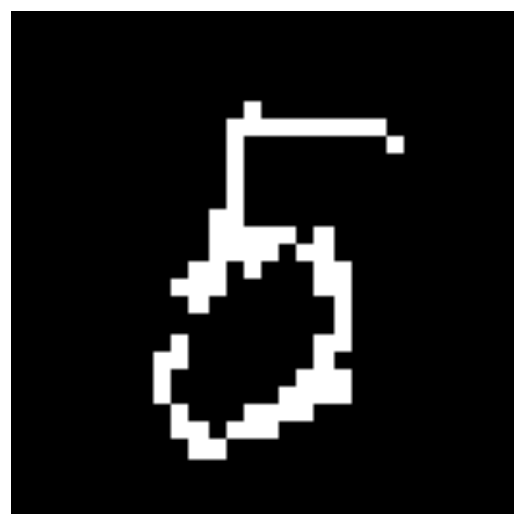} \quad {\footnotesize binary, $\epsilon=0.44$}\\
\includegraphics[align=c,width=0.12\columnwidth]{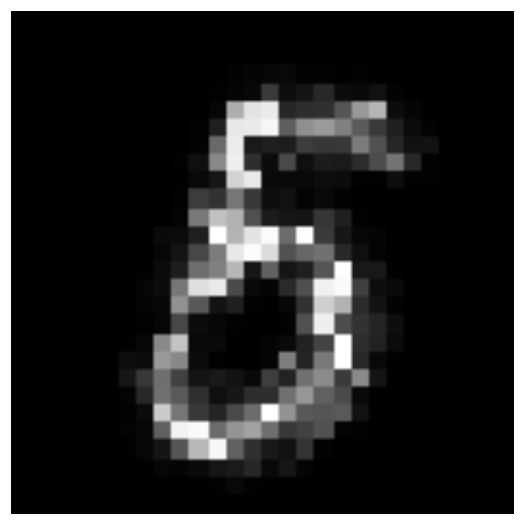} \quad {\footnotesize $\ell_\infty$ robust, $\epsilon=0.78$}\\
\includegraphics[align=c,width=0.12\columnwidth]{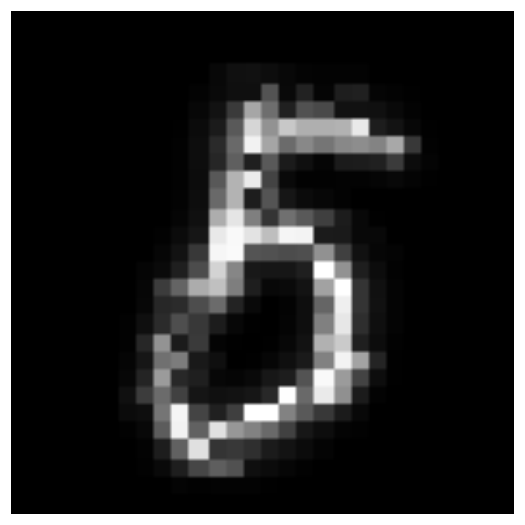} \quad {\footnotesize adv. training, $\epsilon=0.86$}\\
\end{tabular}
\end{tabular}

\vskip -0.1in
\caption{Wasserstein adversarial examples on the MNIST dataset for the four different models. Note that 
the $\ell_\infty$ robust and the adversarially trained models require a much larger $\epsilon$ 
radius for the Wasserstein ball in order to generate an adversarial example. Each model 
classifies the corresponding perturbed example as an 8 instead of a 5,
 except for the first one which classifies the perturbed example as a 6. 
}
\label{fig:mnist_comparison}
\end{center}
\vskip -0.2in
\end{figure}

\begin{figure}[t]
\begin{center}
\includegraphics[scale=0.9]{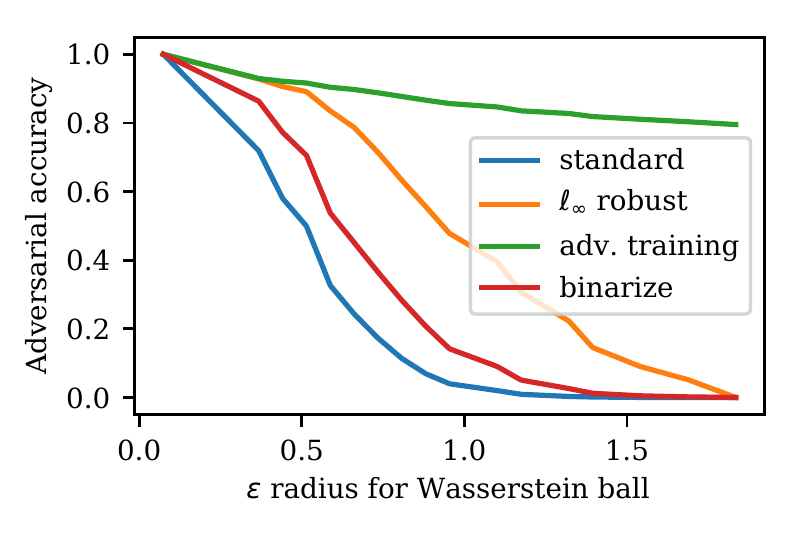}
\vskip -0.2in
\caption{Adversarial accuracy of various models on MNIST 
when attacked by a Wasserstein adversary 
over varying sizes of $\epsilon$-Wasserstein balls. We find that all models 
not trained with adversarial training against this attack eventually achieve 0\% accuracy, 
however we do observe that models trained to be provably robust against $\ell_\infty$ perturbations 
are still somewhat more robust than standard models, or models utilizing binarization as a defense. 
}
\label{fig:mnist_standard_attack}
\end{center}
\end{figure}

For MNIST, we consider a standard model, a model with binarization, a model provably robust 
to $\ell_\infty$ perturbations of at most $\epsilon=0.1$, and an adversarially trained model. 
We provide a visual comparison of the Wasserstein adversarial examples generated 
on each of the four models in Figure \ref{fig:mnist_comparison}. The 
susceptibility of all four models to the Wasserstein attack is plotted in 
Figure \ref{fig:mnist_standard_attack}.

\paragraph{Standard model and binarization} 
For MNIST, despite restricting the transport plan to local 
$5 \times 5$ regions, a standard model is easily attacked by Wasserstein 
adversarial examples. In Figure \ref{fig:mnist_standard_attack}, we see 
that Wasserstein attacks with $\epsilon=0.5$ can successfully attack 
a typical MNIST classifier 50\% of the time, 
which goes up to 94\% for $\epsilon = 1$. A Wasserstein radius of $\epsilon = 0.5$ 
can be intuitively understood as moving 50\% of the pixel mass over by 1 pixel, 
or alternatively moving less than 50\% of the pixel mass more than 1 pixel. 
Furthermore, while preprocessing images with binarization is often seen 
as a way to trivialize adversarial examples on MNIST,
we find that it performs only marginally better than the standard model
against Wasserstein perturbations. 

\paragraph{$\ell_\infty$ robust model}
We also run the attack on the model trained by \citet{wong2018scaling}, which is  
guaranteed to be provably robust against $\ell_\infty$ perturbations with 
$\epsilon \leq 0.1$. 
While not specifically trained against Wasserstein 
perturbations, in Figure \ref{fig:mnist_standard_attack} 
we find that it is substantially more robust than either the standard 
or the binarized model, requiring a significantly larger $\epsilon$ 
to have the same attack success rate. 

\paragraph{Adversarial training}
Finally, we apply this attack as an inner procedure within an adversarial 
training framework for MNIST. To save on computation, 
during training we adopt a weaker adversary and use only 50 
iterations of projected gradient descent. We also let 
$\epsilon$ grow within a range and train on the first adversarial example found 
(essentially a budget version of the attack used at test time). Specific details 
regarding this $\epsilon$ schedule and also the learning parameters used 
can be found in Appendix \ref{app:mnist}. 
We find that the adversarially trained model is empirically the most well defended 
against this attack of all four models, and cannot be attacked down to 0\% accuracy (Figure \ref{fig:mnist_standard_attack}).

\subsection{CIFAR10}
\begin{figure}[t]
\begin{center}
\includegraphics[scale=0.9]{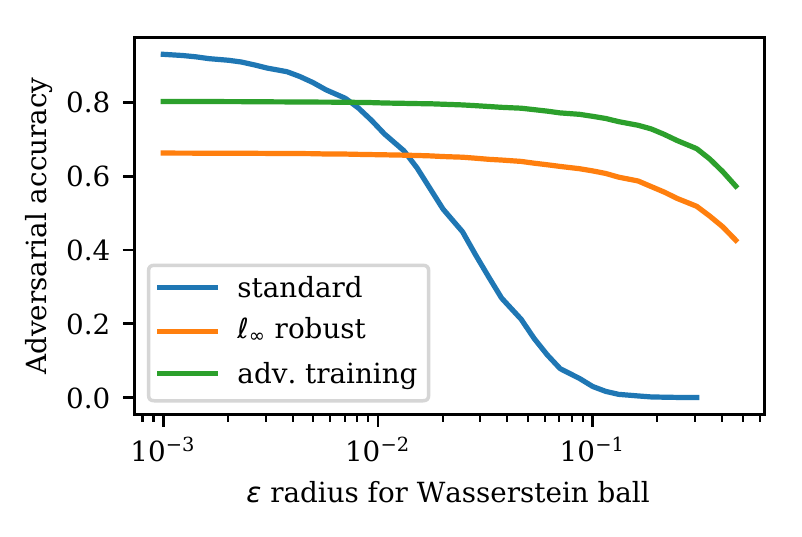}
\vskip -0.2in
\caption{Adversarial accuracy of various models on the CIFAR10 dataset when attacked 
by a Wasserstein adversary. We find that the model trained to be provably robust against $\ell_\infty$ 
perturbations is not as robust as adversarial training against a Wasserstein adversary. 
}
\label{fig:resnet_standard_attack}
\end{center}
\end{figure}

For CIFAR10, we consider a standard model, a model provably robust to $\ell_\infty$ perturbations 
of at most $\epsilon = 2/255$, and an adversarially trained model. We plot 
the susceptibility of each model to the Wasserstein attack in Figure \ref{fig:resnet_standard_attack}.



\begin{figure}[t]
\begin{center}
\setlength{\tabcolsep}{0pt}
\begin{tabular}{crlcccccrlc}
\rotatebox[origin=c]{90}{\footnotesize plane\vphantom{pl}}&
\includegraphics[align=c,width=0.125\columnwidth]{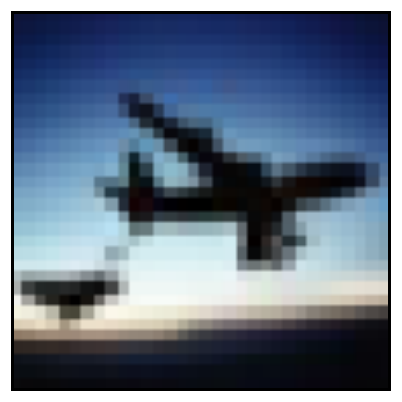}&\hspace{-3px}
\includegraphics[align=c,width=0.125\columnwidth]{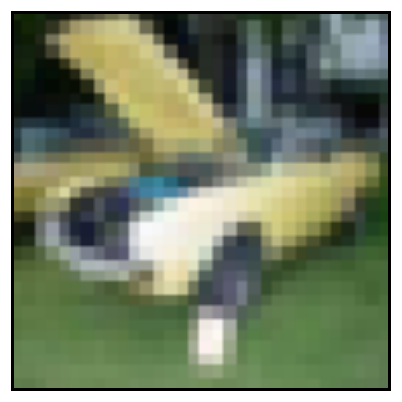}&
\rotatebox[origin=c]{270}{\footnotesize car\vphantom{pl}} 
&  & 
\includegraphics[align=c,width=0.125\columnwidth]{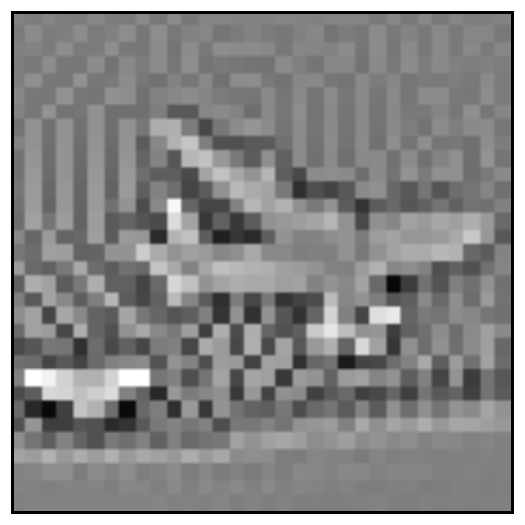}\hspace{-3px}
\includegraphics[align=c,width=0.125\columnwidth]{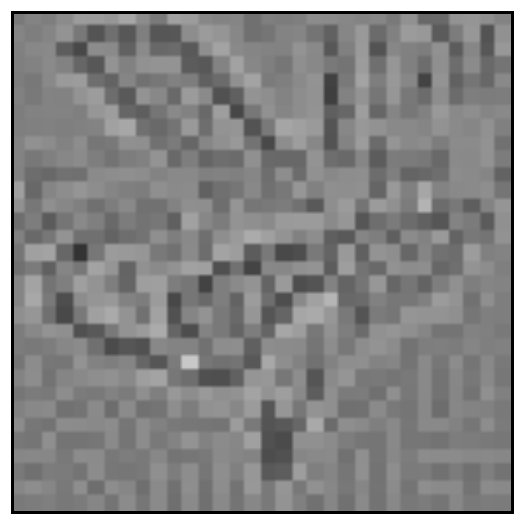} &  & 
\rotatebox[origin=c]{90}{\footnotesize bird\vphantom{pl}}&
\includegraphics[align=c,width=0.125\columnwidth]{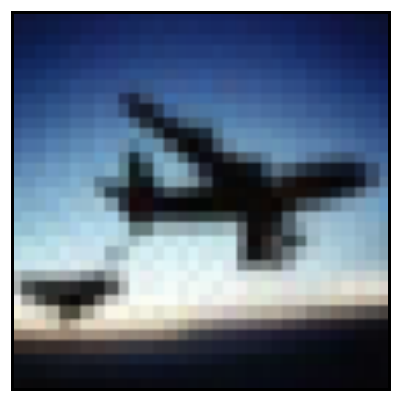}&\hspace{-3px}
\includegraphics[align=c,width=0.125\columnwidth]{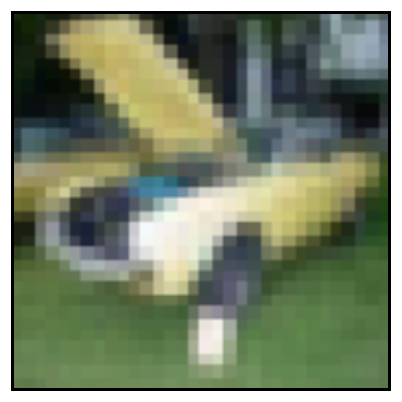}&
\rotatebox[origin=c]{270}{\footnotesize deer\vphantom{pl}} \\
\rotatebox[origin=c]{90}{\footnotesize bird\vphantom{pl}}&
\includegraphics[align=c,width=0.125\columnwidth]{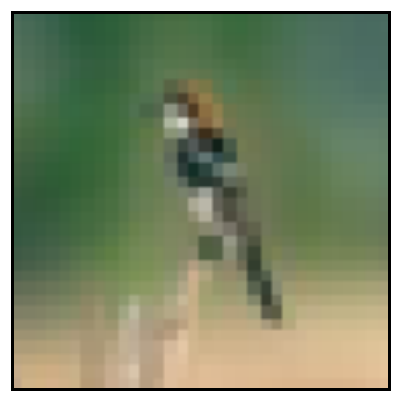}&\hspace{-3px}
\includegraphics[align=c,width=0.125\columnwidth]{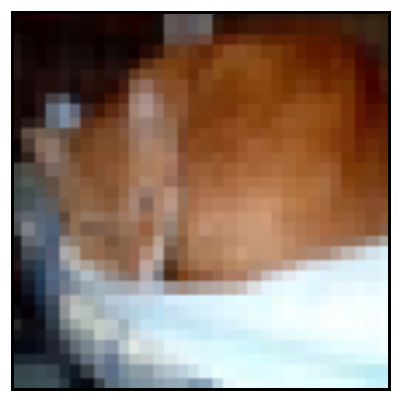}&
\rotatebox[origin=c]{270}{\footnotesize cat\vphantom{pl}} 
&  & 
\includegraphics[align=c,width=0.125\columnwidth]{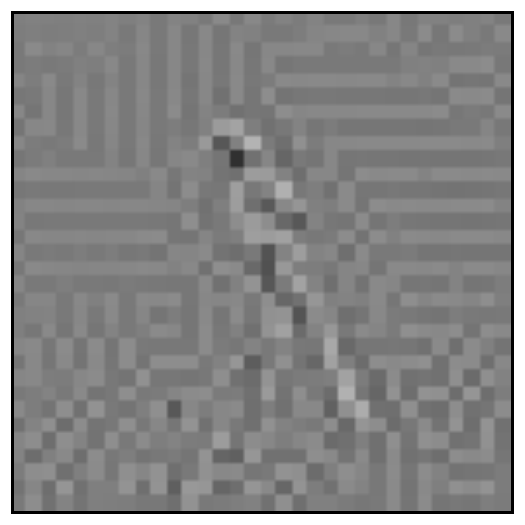}\hspace{-3px}
\includegraphics[align=c,width=0.125\columnwidth]{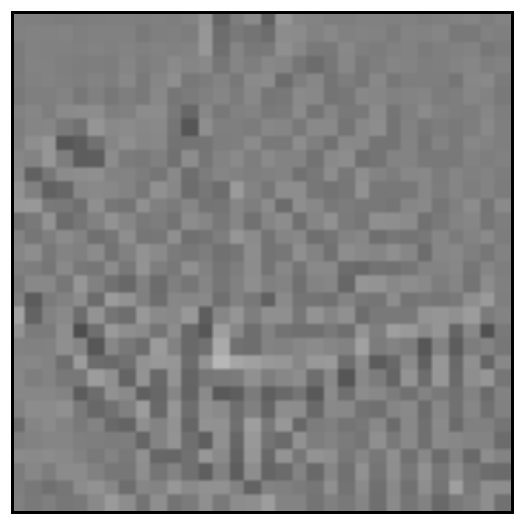} &  & 
\rotatebox[origin=c]{90}{\footnotesize deer\vphantom{pl}}&
\includegraphics[align=c,width=0.125\columnwidth]{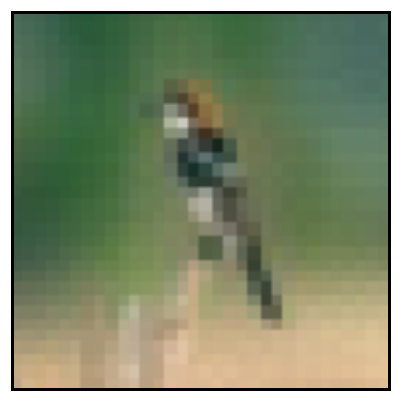}&\hspace{-3px}
\includegraphics[align=c,width=0.125\columnwidth]{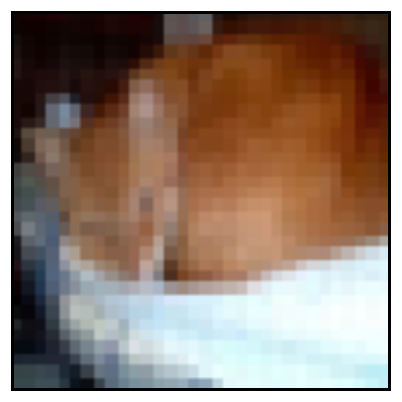}&
\rotatebox[origin=c]{270}{\footnotesize dog\vphantom{pl}} \\
\rotatebox[origin=c]{90}{\footnotesize deer\vphantom{pl}}&
\includegraphics[align=c,width=0.125\columnwidth]{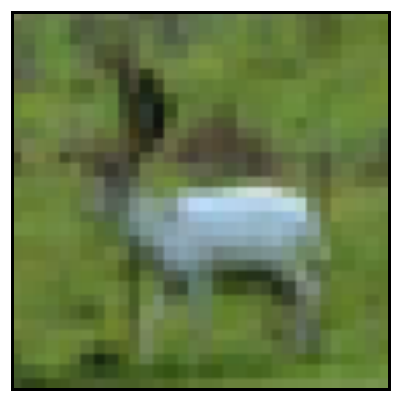}&\hspace{-3px}
\includegraphics[align=c,width=0.125\columnwidth]{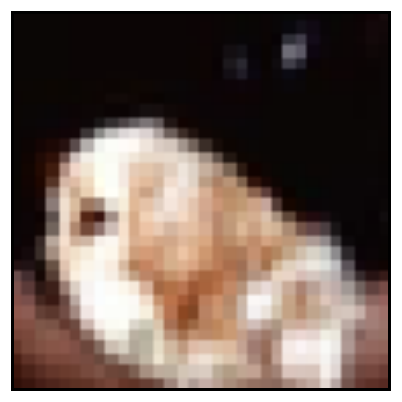}&
\rotatebox[origin=c]{270}{\footnotesize dog\vphantom{pl}} 
& \;+\; & 
\includegraphics[align=c,width=0.125\columnwidth]{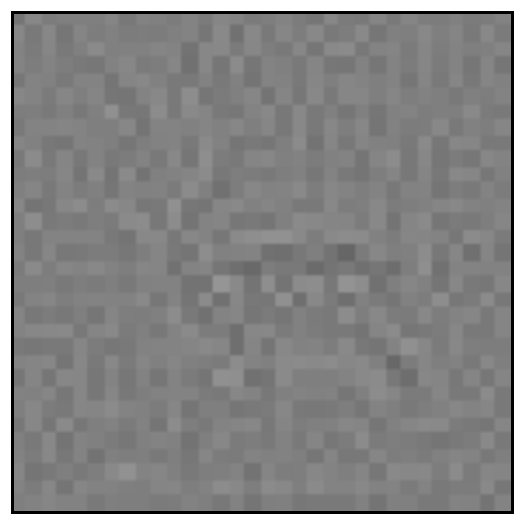}\hspace{-3px}
\includegraphics[align=c,width=0.125\columnwidth]{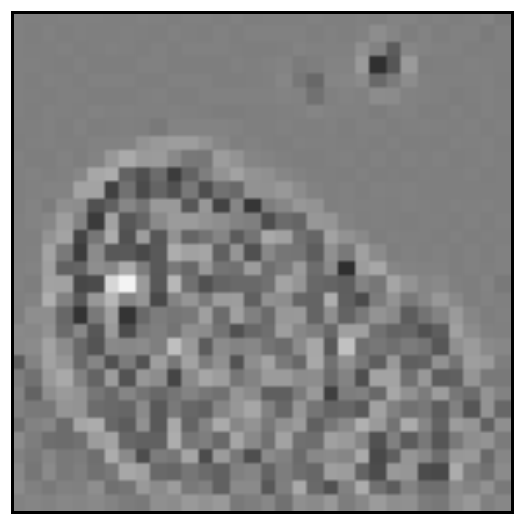} & \;=\; & 
\rotatebox[origin=c]{90}{\footnotesize horse\vphantom{pl}}&
\includegraphics[align=c,width=0.125\columnwidth]{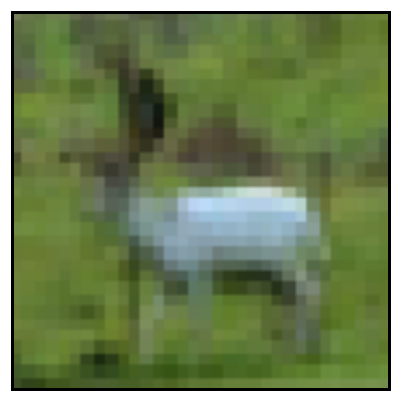}&\hspace{-3px}
\includegraphics[align=c,width=0.125\columnwidth]{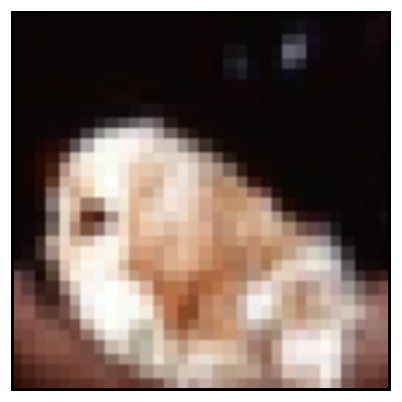}&
\rotatebox[origin=c]{270}{\footnotesize cat\vphantom{pl}} \\
\rotatebox[origin=c]{90}{\footnotesize frog\vphantom{pl}}&
\includegraphics[align=c,width=0.125\columnwidth]{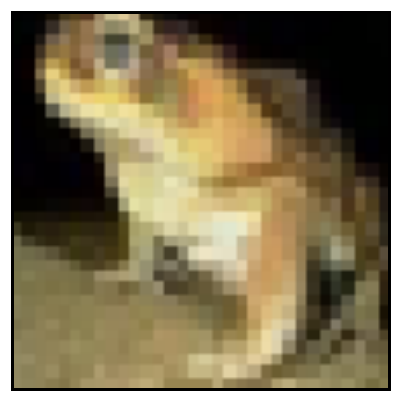}&\hspace{-3px}
\includegraphics[align=c,width=0.125\columnwidth]{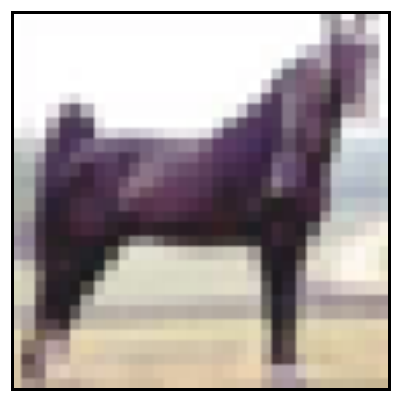}&
\rotatebox[origin=c]{270}{\footnotesize horse\vphantom{pl}} 
&  & 
\includegraphics[align=c,width=0.125\columnwidth]{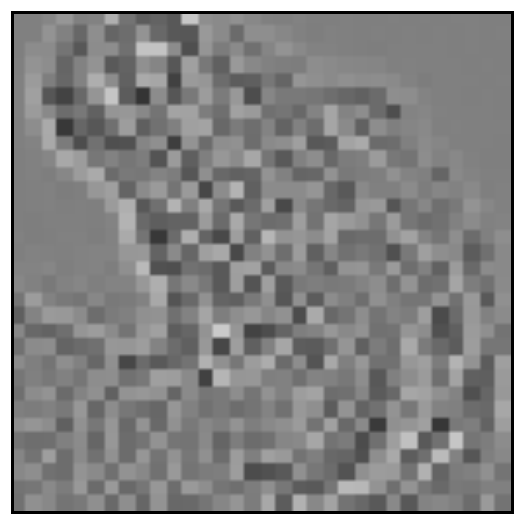}\hspace{-3px}
\includegraphics[align=c,width=0.125\columnwidth]{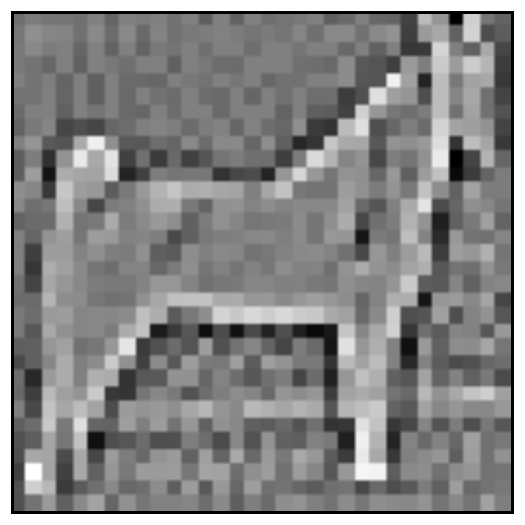} &  & 
\rotatebox[origin=c]{90}{\footnotesize bird\vphantom{pl}}&
\includegraphics[align=c,width=0.125\columnwidth]{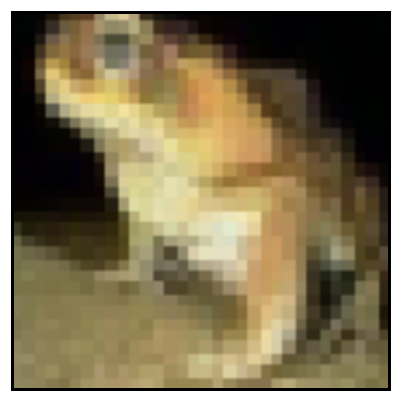}&\hspace{-3px}
\includegraphics[align=c,width=0.125\columnwidth]{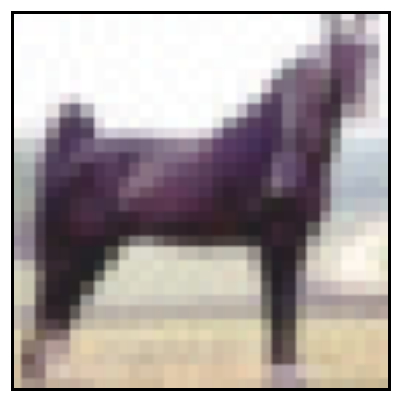}&
\rotatebox[origin=c]{270}{\footnotesize dog\vphantom{pl}} \\
\rotatebox[origin=c]{90}{\footnotesize ship\vphantom{pl}}&
\includegraphics[align=c,width=0.125\columnwidth]{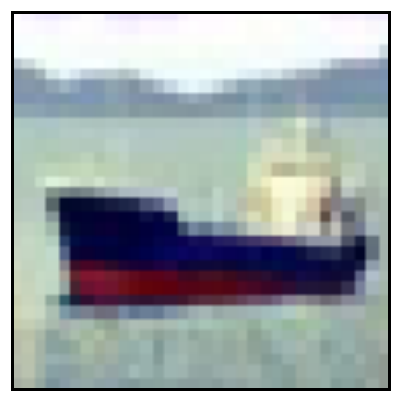}&\hspace{-3px}
\includegraphics[align=c,width=0.125\columnwidth]{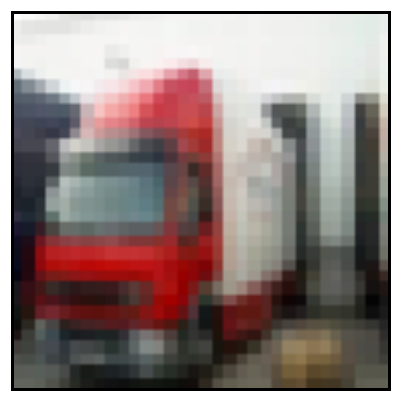}&
\rotatebox[origin=c]{270}{\footnotesize truck\vphantom{pl}} 
&  & 
\includegraphics[align=c,width=0.125\columnwidth]{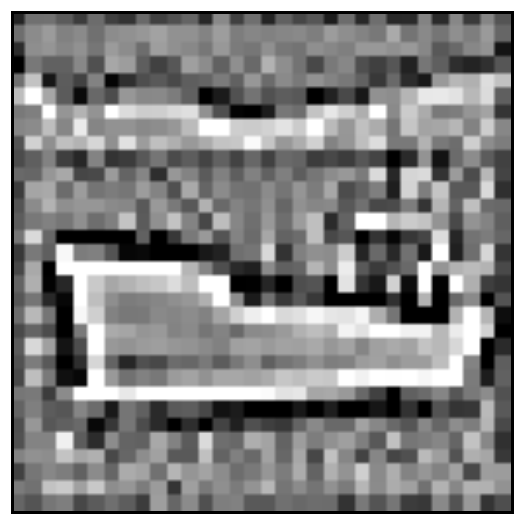}\hspace{-3px}
\includegraphics[align=c,width=0.125\columnwidth]{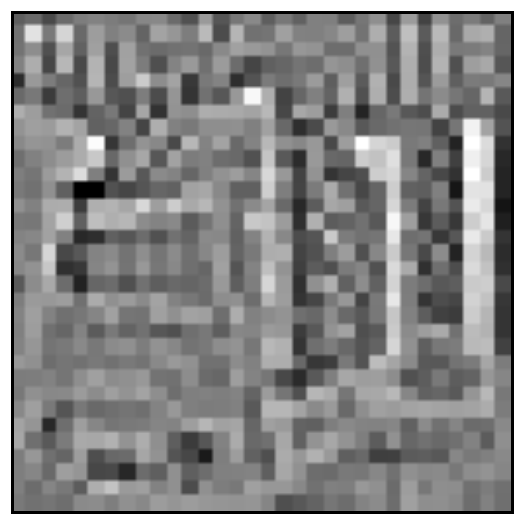} &  & 
\rotatebox[origin=c]{90}{\footnotesize plane\vphantom{pl}}&
\includegraphics[align=c,width=0.125\columnwidth]{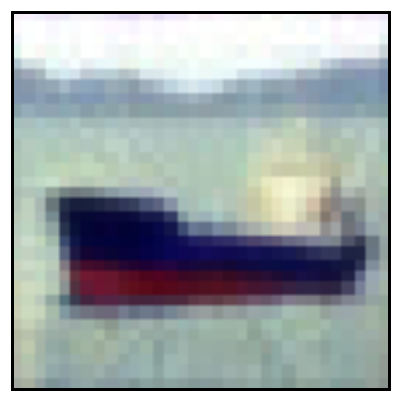}&\hspace{-3px}
\includegraphics[align=c,width=0.125\columnwidth]{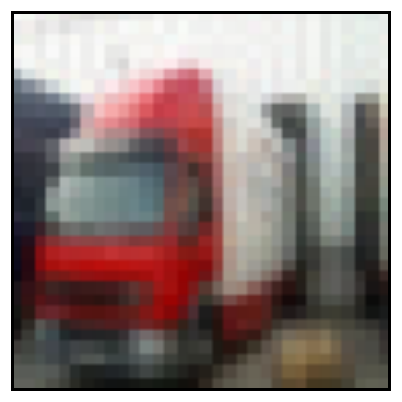}&
\rotatebox[origin=c]{270}{\footnotesize car\vphantom{pl}} \\
\end{tabular}

\vskip -0.1in
\caption{Wasserstein adversarial examples for CIFAR10 on a typical ResNet18 for all 10 classes. 
The perturbations here represents the total change across all three channels, where total 
change is plotted within the range  $\pm 0.165$ (the maximum total change observed in a single pixel) for images scaled to [0,1]. 
}
\label{fig:perturbationdiff}
\end{center}
\vskip -0.2in
\end{figure}

\paragraph{Standard model}
We find that for a standard ResNet18 CIFAR10 classifier, a perturbation radius of as little as 
$0.01$ is enough to misclassify 25\% of the examples, while a radius of $0.1$ 
is enough to fool the classifier 97\% of the time (Figure \ref{fig:resnet_standard_attack}). 
Despite being such a small $\epsilon$, we see in Figure \ref{fig:perturbationdiff} that the 
structure of the perturbations still reflect the actual content of the images, 
though certain classes require larger magnitudes of change than others. 

\paragraph{$\ell_\infty$ robust model}
We further empirically evaluate the attack on a model that was trained to 
be provably robust against $\ell_\infty$ perturbations. We use the models weights 
from \citet{wong2018scaling}, which are trained to be provably robust against $\ell_\infty$ 
perturbations of at most $\epsilon = 2/255$. 
Further note that this CIFAR10 model actually is 
a smaller ResNet than the ResNet18 architecture considered in this paper, 
and consists of 4 residual blocks with 16, 16, 32, and 64 filters.  
Nonetheless, we find that while the model suffers from poor nominal accuracy 
(achieving only 66\% accuracy on unperturbed examples as noted in Table \ref{tab:nominal}), 
the robustness against $\ell_\infty$ attacks remarkably seems to transfer quite well to 
robustness against Wasserstein attacks in the CIFAR10 setting, achieving 61\% adversarial 
accuracy for $\epsilon=0.1$ in comparison to 3\% for the standard model.

\paragraph{Adversarial training}
To perform adversarial training for CIFAR10, we use a similar scheme to that used 
for MNIST: we adopt a weaker adversary that uses only 50 iterations of projected gradient descent during training and allow $\epsilon$ to grow within a range (specific details can be found in 
Appendix \ref{app:cifar10}). 
We find that adversarial training here is also able to defend against 
this attack, and at the same threshold 
of $\epsilon = 0.1$, we find that the adversarial accuracy has been improved 
from  3\% to 76\%. 



\subsection{Provable Defenses against Wasserstein Perturbations}
Lastly, we present some analysis on how this attack fits 
into the context of provable defenses, along with a negative 
result demonstrating a fundamental gap that needs to be solved. 
The Wasserstein attack 
can be naturally incorporated into duality based defenses: 
\citet{wong2018scaling} show that to use their certificates 
to defend against other inputs, 
one only needs to solve the following optimization problem: 
\begin{equation}
\max_{x \in B(x,\epsilon)} -x^Ty
\label{eq:provable}
\end{equation}
for some constant $y$ and for some perturbation region $B(x,\epsilon)$ 
(a similar approach can be taken to adapt the dual verification from \citet{dvijotham18}). 
For the Wasserstein ball, this is highly similar to 
the problem of projecting onto the Wasserstein 
ball from Equation \eqref{eq:projection_reg}, with 
a linear objective instead 
of a quadratic objective and fewer variables. In fact, a 
Sinkhorn-like algorithm can be derived to solve this problem, which 
ends up being a simplified version of Algorithm \ref{alg:projected_sinkhorn} 
(this is shown in Appendix \ref{app:provable}). 

However, there is a 
fundamental obstacle towards generating provable certificates
against Wasserstein attacks: 
these defenses (and many other, non-duality based approaches) depend 
heavily on propagating interval bounds from the input space 
through the network, in order to efficiently bound the output of 
ReLU units. This concept is inherently at odds with the notion of 
Wasserstein distance: a ``small'' Wasserstein ball 
can use a low-cost transport plan to move all the mass at a single pixel to 
its neighbors, or vice versa. As a result, when converting a 
Wasserstein ball to interval constraints, the interval bounds 
immediately become vacuous: each individual pixel can attain their 
minimum or maximum value under some $\epsilon$ cost transport plan. 
In order to guarantee robustness against Wasserstein adversarial attacks, 
significant progress must be made to overcome this limitation. 


\section{Conclusion}
In this paper, we have presented a new, general threat model for adversarial 
examples based on the Wasserstein distance, a metric that captures a 
kind of perturbation that is fundamentally different from traditional $\ell_p$ 
perturbations. To generate these examples, we derived an algorithm for 
fast, approximate projection onto the Wasserstein ball that can use local transport plans
for even more speedup on images.  We successfully attacked standard networks, showing 
that these adversarial examples are structurally perturbed according to the 
content of the image, and demonstrated the empirical effectiveness of adversarial training. 
Finally, we observed that networks trained to be provably robust against $\ell_\infty$ attacks 
are more robust than the standard networks against Wasserstein attacks, however 
we show that the current state of provable defenses is insufficient to 
directly apply to the Wasserstein ball due to their reliance on interval bounds. 

We believe overcoming this roadblock is crucial to the development of 
verifiers or provable defenses against not just the Wasserstein attack, but also 
to improve the robustness of classifiers against other attacks that do not naturally 
convert to interval bounds (e.g. $\ell_0$ or $\ell_1$ attacks). 
Whether we can develop efficient verification or provable training methods that do not 
rely on interval bounds remains an open question. 

Perhaps the most natural future direction for this work is to begin to understand 
the properties of Wasserstein adversarial examples and what we can do to mitigate them, 
even if only at a heuristic level. However, at the end of the day, the Wasserstein threat model 
defines just one example of a convex region capturing structure 
that is different from $\ell_p$ balls. 
By no means have we characterized all reasonable adversarial perturbations, 
and so a significant gap remains in determining how to rigorously 
define general classes of adversarial examples that can characterize phenomena 
different from the $\ell_p$ and Wasserstein balls. 

Finally, although we focused primarily on adversarial examples in this work, the 
method of projecting onto Wasserstein balls may be applicable outside of 
deep learning. Projection operators play a major role in optimization algorithms 
beyond projected gradient descent (e.g. ADMM and alternating projections). Perhaps even 
more generally, the techniques in this paper could be used to derive Sinkhorn-like 
algorithms for classes of problems that consider Wasserstein constrained variables. 

\bibliography{projected_sinkhorn}

\begin{thebibliography}{25}
\providecommand{\natexlab}[1]{#1}
\providecommand{\url}[1]{\texttt{#1}}
\expandafter\ifx\csname urlstyle\endcsname\relax
  \providecommand{\doi}[1]{doi: #1}\else
  \providecommand{\doi}{doi: \begingroup \urlstyle{rm}\Url}\fi

\bibitem[Altschuler et~al.(2017)Altschuler, Weed, and
  Rigollet]{altschuler2017near}
Altschuler, J., Weed, J., and Rigollet, P.
\newblock Near-linear time approximation algorithms for optimal transport via
  sinkhorn iteration.
\newblock In \emph{Advances in Neural Information Processing Systems}, pp.\
  1964--1974, 2017.

\bibitem[Athalye et~al.(2018{\natexlab{a}})Athalye, Carlini, and
  Wagner]{obfuscated-gradients}
Athalye, A., Carlini, N., and Wagner, D.
\newblock Obfuscated gradients give a false sense of security: Circumventing
  defenses to adversarial examples.
\newblock In \emph{Proceedings of the 35th International Conference on Machine
  Learning, {ICML} 2018}, July 2018{\natexlab{a}}.
\newblock URL \url{https://arxiv.org/abs/1802.00420}.

\bibitem[Athalye et~al.(2018{\natexlab{b}})Athalye, Engstrom, Ilyas, and
  Kwok]{pmlr-v80-athalye18b}
Athalye, A., Engstrom, L., Ilyas, A., and Kwok, K.
\newblock Synthesizing robust adversarial examples.
\newblock In Dy, J. and Krause, A. (eds.), \emph{Proceedings of the 35th
  International Conference on Machine Learning}, volume~80 of \emph{Proceedings
  of Machine Learning Research}, pp.\  284--293, Stockholmsmässan, Stockholm
  Sweden, 10--15 Jul 2018{\natexlab{b}}. PMLR.
\newblock URL \url{http://proceedings.mlr.press/v80/athalye18b.html}.

\bibitem[Carlini \& Wagner(2017)Carlini and Wagner]{carlini2017towards}
Carlini, N. and Wagner, D.
\newblock Towards evaluating the robustness of neural networks.
\newblock In \emph{Security and Privacy (SP), 2017 IEEE Symposium on}, pp.\
  39--57. IEEE, 2017.

\bibitem[Croce et~al.(2018)Croce, Andriushchenko, and Hein]{croce2018provable}
Croce, F., Andriushchenko, M., and Hein, M.
\newblock Provable robustness of relu networks via maximization of linear
  regions.
\newblock \emph{CoRR}, abs/1810.07481, 2018.
\newblock URL \url{http://arxiv.org/abs/1810.07481}.

\bibitem[Cuturi(2013)]{cuturi2013sinkhorn}
Cuturi, M.
\newblock Sinkhorn distances: Lightspeed computation of optimal transport.
\newblock In Burges, C. J.~C., Bottou, L., Welling, M., Ghahramani, Z., and
  Weinberger, K.~Q. (eds.), \emph{Advances in Neural Information Processing
  Systems 26}, pp.\  2292--2300. Curran Associates, Inc., 2013.
\newblock URL
  \url{http://papers.nips.cc/paper/4927-sinkhorn-distances-lightspeed-computation-of-optimal-transport.pdf}.

\bibitem[Dvijotham et~al.(2018)Dvijotham, Stanforth, Gowal, Mann, and
  Kohli]{dvijotham18}
Dvijotham, K., Stanforth, R., Gowal, S., Mann, T., and Kohli, P.
\newblock A dual approach to scalable verification of deep networks.
\newblock In \emph{Proceedings of the Thirty-Fourth Conference Annual
  Conference on Uncertainty in Artificial Intelligence (UAI-18)}, pp.\
  162--171, Corvallis, Oregon, 2018. AUAI Press.

\bibitem[Engstrom et~al.(2017)Engstrom, Tran, Tsipras, Schmidt, and
  Madry]{engstrom2017rotation}
Engstrom, L., Tran, B., Tsipras, D., Schmidt, L., and Madry, A.
\newblock A rotation and a translation suffice: Fooling cnns with simple
  transformations.
\newblock \emph{arXiv preprint arXiv:1712.02779}, 2017.

\bibitem[Eykholt et~al.(2018)Eykholt, Evtimov, Fernandes, Li, Rahmati, Xiao,
  Prakash, Kohno, and Song]{eykholt2018robust}
Eykholt, K., Evtimov, I., Fernandes, E., Li, B., Rahmati, A., Xiao, C.,
  Prakash, A., Kohno, T., and Song, D.
\newblock Robust physical-world attacks on deep learning visual classification.
\newblock In \emph{Proceedings of the IEEE Conference on Computer Vision and
  Pattern Recognition}, pp.\  1625--1634, 2018.

\bibitem[Goodfellow et~al.(2015)Goodfellow, Shlens, and
  Szegedy]{goodfellow2015explaining}
Goodfellow, I., Shlens, J., and Szegedy, C.
\newblock Explaining and harnessing adversarial examples.
\newblock In \emph{International Conference on Learning Representations}, 2015.
\newblock URL \url{http://arxiv.org/abs/1412.6572}.

\bibitem[Gowal et~al.(2018)Gowal, Dvijotham, Stanforth, Bunel, Qin, Uesato,
  Arandjelovic, Mann, and Kohli]{gowal2018interval}
Gowal, S., Dvijotham, K., Stanforth, R., Bunel, R., Qin, C., Uesato, J.,
  Arandjelovic, R., Mann, T.~A., and Kohli, P.
\newblock On the effectiveness of interval bound propagation for training
  verifiably robust models.
\newblock \emph{CoRR}, abs/1810.12715, 2018.
\newblock URL \url{http://arxiv.org/abs/1810.12715}.

\bibitem[He et~al.(2016)He, Zhang, Ren, and Sun]{he2016deep}
He, K., Zhang, X., Ren, S., and Sun, J.
\newblock Deep residual learning for image recognition.
\newblock In \emph{Proceedings of the IEEE conference on computer vision and
  pattern recognition}, pp.\  770--778, 2016.

\bibitem[Kurakin et~al.(2017)Kurakin, Goodfellow, and
  Bengio]{kurakin2017adversarial}
Kurakin, A., Goodfellow, I., and Bengio, S.
\newblock Adversarial examples in the physical world.
\newblock \emph{ICLR Workshop}, 2017.
\newblock URL \url{https://arxiv.org/abs/1607.02533}.

\bibitem[Lu et~al.(2017)Lu, Sibai, Fabry, and Forsyth]{lu2017no}
Lu, J., Sibai, H., Fabry, E., and Forsyth, D.
\newblock No need to worry about adversarial examples in object detection in
  autonomous vehicles.
\newblock \emph{arXiv preprint arXiv:1707.03501}, 2017.

\bibitem[Madry et~al.(2018)Madry, Makelov, Schmidt, Tsipras, and
  Vladu]{madry2018towards}
Madry, A., Makelov, A., Schmidt, L., Tsipras, D., and Vladu, A.
\newblock Towards deep learning models resistant to adversarial attacks.
\newblock In \emph{International Conference on Learning Representations}, 2018.
\newblock URL \url{https://openreview.net/forum?id=rJzIBfZAb}.

\bibitem[Mirman et~al.(2018)Mirman, Gehr, and Vechev]{mirman2018diff}
Mirman, M., Gehr, T., and Vechev, M.
\newblock Differentiable abstract interpretation for provably robust neural
  networks.
\newblock In \emph{International Conference on Machine Learning (ICML)}, 2018.
\newblock URL
  \url{https://www.icml.cc/Conferences/2018/Schedule?showEvent=2477}.

\bibitem[Papernot et~al.(2016)Papernot, McDaniel, Wu, Jha, and
  Swami]{papernot2016distillation}
Papernot, N., McDaniel, P., Wu, X., Jha, S., and Swami, A.
\newblock Distillation as a defense to adversarial perturbations against deep
  neural networks.
\newblock In \emph{Security and Privacy (SP), 2016 IEEE Symposium on}, pp.\
  582--597. IEEE, 2016.

\bibitem[Raghunathan et~al.(2018)Raghunathan, Steinhardt, and
  Liang]{raghunathan2018semi}
Raghunathan, A., Steinhardt, J., and Liang, P.~S.
\newblock Semidefinite relaxations for certifying robustness to adversarial
  examples.
\newblock In Bengio, S., Wallach, H., Larochelle, H., Grauman, K.,
  Cesa-Bianchi, N., and Garnett, R. (eds.), \emph{Advances in Neural
  Information Processing Systems 31}, pp.\  10900--10910. Curran Associates,
  Inc., 2018.
\newblock URL
  \url{http://papers.nips.cc/paper/8285-semidefinite-relaxations-for-certifying-robustness-to-adversarial-examples.pdf}.

\bibitem[Sharif et~al.(2017)Sharif, Bhagavatula, Bauer, and
  Reiter]{sharif2017adversarial}
Sharif, M., Bhagavatula, S., Bauer, L., and Reiter, M.~K.
\newblock Adversarial generative nets: Neural network attacks on
  state-of-the-art face recognition.
\newblock \emph{arXiv preprint arXiv:1801.00349}, 2017.

\bibitem[Sinha et~al.(2018)Sinha, Namkoong, and Duchi]{sinha2018certifying}
Sinha, A., Namkoong, H., and Duchi, J.
\newblock Certifying some distributional robustness with principled adversarial
  training.
\newblock 2018.

\bibitem[Szegedy et~al.(2014)Szegedy, Zaremba, Sutskever, Bruna, Erhan,
  Goodfellow, and Fergus]{szegedy2014intriguing}
Szegedy, C., Zaremba, W., Sutskever, I., Bruna, J., Erhan, D., Goodfellow, I.,
  and Fergus, R.
\newblock Intriguing properties of neural networks.
\newblock In \emph{International Conference on Learning Representations}, 2014.
\newblock URL \url{http://arxiv.org/abs/1312.6199}.

\bibitem[Tjeng et~al.(2019)Tjeng, Xiao, and Tedrake]{tjeng2018evaluating}
Tjeng, V., Xiao, K.~Y., and Tedrake, R.
\newblock Evaluating robustness of neural networks with mixed integer
  programming.
\newblock In \emph{International Conference on Learning Representations}, 2019.
\newblock URL \url{https://openreview.net/forum?id=HyGIdiRqtm}.

\bibitem[Wong \& Kolter(2018)Wong and Kolter]{wong2018provable}
Wong, E. and Kolter, Z.
\newblock Provable defenses against adversarial examples via the convex outer
  adversarial polytope.
\newblock In \emph{International Conference on Machine Learning}, pp.\
  5283--5292, 2018.

\bibitem[Wong et~al.(2018)Wong, Schmidt, Metzen, and Kolter]{wong2018scaling}
Wong, E., Schmidt, F., Metzen, J.~H., and Kolter, J.~Z.
\newblock Scaling provable adversarial defenses.
\newblock In Bengio, S., Wallach, H., Larochelle, H., Grauman, K.,
  Cesa-Bianchi, N., and Garnett, R. (eds.), \emph{Advances in Neural
  Information Processing Systems 31}, pp.\  8410--8419. Curran Associates,
  Inc., 2018.
\newblock URL
  \url{http://papers.nips.cc/paper/8060-scaling-provable-adversarial-defenses.pdf}.

\bibitem[Xiao et~al.(2019)Xiao, Tjeng, Shafiullah, and Madry]{xiao2018training}
Xiao, K.~Y., Tjeng, V., Shafiullah, N. M.~M., and Madry, A.
\newblock Training for faster adversarial robustness verification via inducing
  re{LU} stability.
\newblock In \emph{International Conference on Learning Representations}, 2019.
\newblock URL \url{https://openreview.net/forum?id=BJfIVjAcKm}.

\end{thebibliography}
\bibliographystyle{icml2019}

\clearpage
\appendix
\section{Projected Sinkhorn derivation}
\subsection{Proof of Lemma \ref{lem:dual}}
\label{app:dual}
\duallemma
\begin{proof}
For convenience, we multiply the objective by $\lambda$ and solve this problem instead: 
 \begin{equation}
\begin{split}
\minimize_{z\in\mathbb R^n_+, \Pi\in \mathbb{R}^{n\times n}_+} &\;\; \frac{\lambda}{2}\|w-z\|_2^2 + \sum_{ij}\Pi_{ij}\log(\Pi_{ij})\\
\subjectto \;\; & \Pi1 = x\\
& \Pi^T1 = z\\
& \langle \Pi,C\rangle \leq \epsilon. 
\end{split}
\end{equation}
Introducing dual variables $(\alpha, \beta, \psi)$ where $\psi \geq 0$, the Lagrangian is 
\begin{equation}
\begin{split}
&L(z, \Pi, \alpha, \beta, \psi) \\
= &\frac{\lambda}{2}\|w-z\|_2^2 + \sum_{ij}\Pi_{ij}\log(\Pi_{ij})+ \psi (\langle \Pi,C\rangle - \epsilon) \\
& + \alpha^T(x - \Pi1) + \beta^T(z - \Pi^T1). 
\end{split}
\end{equation}
The KKT optimality conditions are now
\begin{equation}
\begin{split}
\frac{\partial L}{\partial \Pi_{ij}} &= \psi C_{ij} + (1 + \log(\Pi_{ij})) - \alpha_i - \beta_j = 0\\
\frac{\partial L}{\partial z_j} &= \lambda(z_j-w_j) + \beta_j = 0
\end{split}
\end{equation}
so at optimality, we must have
\begin{equation}
\begin{split}
\Pi_{ij} &= \exp(\alpha_i)\exp(-\psi C_{ij}-1)\exp(\beta_j)\\
z  &= -\frac{\beta}{\lambda} + w
\end{split}
\label{eq:kkt}
\end{equation}
Plugging in the optimality conditions, we get
\begin{equation}
\begin{split}
& L(z^*,\Pi^*, \alpha, \beta, \psi) \\
=& -\frac{1}{2\lambda}\|\beta\|_2^2 - \psi \epsilon + \alpha^Tx + \beta^Tw \\
& - \sum_{ij}\exp(\alpha_i)\exp( - \psi C_{ij} - 1)\exp( \beta_j)\\
=& g(\alpha, \beta, \psi)
\end{split}
\end{equation}
so the dual problem is to maximize $g$ over $\alpha, \beta, \psi \geq 0$.
\end{proof}

\subsection{Proof of Lemma \ref{lem:dualtoprimal}}
\label{app:dualtoprimal}
\dualtoprimal
\begin{proof}
These equations follow directly from the KKT optimality conditions from Equation \eqref{eq:kkt}. 
\end{proof}

\subsection{Algorithm derivation and interpretation}
\label{app:dualexplain}
\paragraph{Derivation}
To derive the algorithm, note that since this is a strictly convex problem 
to get the $\alpha$ and $\beta$ iterates we solve for setting the gradient to 0. 
The derivative with respect to $\alpha$ is
\begin{equation}
\frac{\partial g}{\partial \alpha_i} = x_i - \exp(\alpha_i) \sum_j  \exp(-\psi C_{ij} - 1) \exp(\beta_j)
\label{eq:deriv_alpha}
\end{equation}
and so setting this to 0 and solving for $\alpha_i$ gives the $\alpha$ iterate. 
The derivative with respect to $\beta$ is 
\begin{equation}
\frac{\partial g}{\partial \beta_j} = -\frac{1}{\lambda}\beta + w -  \exp(\beta_j)\sum_i  \exp(\alpha_i)\exp(-\psi C_{ij} - 1) 
\label{eq:deriv_beta}
\end{equation}
and setting this to 0 and solving for $\beta_j$ gives the $\beta$ iterate (this step can be done 
using a symbolic solver, we used Mathematica). Lastly, the $\psi$ updates are straightforward 
scalar calculations of the derivative and second derivative. 

\paragraph{Interpretation}
Recall from the transformation of dual to primal variables from Lemma \ref{lem:dualtoprimal}. 
To see how the Projected Sinkhorn iteration is a (modified) matrix scaling algorithm, 
we can interpret these quantities before optimality as primal iterates. Namely, at 
each iteration $t$, let
\begin{equation}
\begin{split}
z_i^{(t)} &= w_i - \beta_i^{(t)}/\lambda\\
\Pi_{ij}^{(t)} &= \exp(\alpha^{(t)})\exp(-\psi^{(t)} C_{ij} - 1)\exp(\beta^{(t)})
\end{split}
\end{equation}
Then, since the $\alpha$ and $\beta$ steps are equivalent to setting 
Equations \eqref{eq:deriv_alpha} and \eqref{eq:deriv_beta} to 0, we know that after an update for 
$\alpha^{(t)}$, we have that 
\begin{equation}
x_i = \sum_j \Pi_{ij}^{(t)}
\end{equation}
so the $\alpha$ step rescales the transport matrix to sum to $x$. Similarly, 
after an update for $\beta^{(t)}$, we have that
\begin{equation}
z_i^{(t)} = \sum_i \Pi_{ij}^{(t)}
\end{equation}
which is a rescaling of the transport matrix to sum to the projected value. 
Lastly, the numerator of the $\psi^{(t)}$ step can be rewritten as 
\begin{equation}
\psi^{(t+1)} = \psi^{(t)} + t\cdot \frac{\langle \Pi^{(t)}, C\rangle -\epsilon}{\langle \Pi^{(t)}, C\cdot C \rangle }
\end{equation}
as a simple adjustment based on whether the current transport plan $\Pi^{(t)}$ is above or 
below the maximum threshold $\epsilon$. 


\section{Experimental setup}
\subsection{MNIST}
\label{app:mnist}
\paragraph{Adaptive $\epsilon$}
During adversarial training for MNIST, we adopt an adaptive $\epsilon$ scheme to avoid 
selecting a specific $\epsilon$. Specifically, to find an adversarial example, we first 
let $\epsilon=0.1$ on the first iteration of projected gradient descent, and increase it 
by a factor of $1.4$ every 5 iterations. We terminate the projected gradient descent algorithm 
when either an adversarial example is found, or when 50 iterations have passed, allowing 
$\epsilon$ to take on values in the range $[0.1, 2.1]$

\paragraph{Optimizer hyperparameters}
To update the model weights during adversarial training, we use the SGD optimizer with 
0.9 momentum and 0.0005 weight decay, and batch sizes of 128. We begin with a 
learning rate of 0.1, reduce it to 0.01 after 10 epochs. 

\subsection{CIFAR10}
\label{app:cifar10}
\paragraph{Adaptive $\epsilon$}
We also use an adaptive $\epsilon$ scheme for adversarial training in CIFAR10. Specifically, 
we let $\epsilon=0.01$ on the first iteration of projected gradient descent, and increase it by a 
factor of 1.5 every 5 iterations. Similar to MNIST, we terminate the projected gradient descent algorithm 
when either an adversarial example is found, or 50 iterations have passed, allowing $\epsilon$ to take on 
values in the range $[0.01, 0.38]$. 
\paragraph{Optimizer hyperparameters}
Similar to MNIST, to update the model weights, we use the SGD optimizer with 0.9 momentum and 0.0005 
weight decay, and batch sizes of 128. The learning rate is also the same as in MNIST, starting at 0.1, 
and reducing to 0.01 after 10 epochs. 
\subsection{Motivation for adaptive $\epsilon$} 
A commonly asked question of models trained to be robust against adversarial examples is 
``what if the adversary has a perturbation budget of 
$\epsilon + \delta$ instead of $\epsilon$?'' This is referring to a
``robustness cliff,'' where a model trained against an $\epsilon$ strong 
adversary has a sharp drop in robustness when attacked by an 
adversary with a slightly larger budget. 
To address this, we advocate for the slightly modified version of typical adversarial training 
used in this work: 
rather than picking a fixed $\epsilon$ and running projected gradient descent, 
we instead allow for an adversarial to have a range of 
$\epsilon \in [\epsilon_{min}, \epsilon_{max}]$. 
To do this, we begin with $\epsilon = \epsilon_{\min}$, and then gradually increase 
it by a multiplicative factor $\gamma$ until either an adversarial example is found 
or until $\epsilon_{max}$ is reached. 
While similar ideas have been used before for evaluating model 
robustness, 
we specifically advocate for using this schema \emph{during adversarial training}. 
This has the advantage of extending robustness of the classifier beyond 
a single $\epsilon$ threshold, allowing a model to achieve a potentially 
higher robustness threshold while not being significantly harmed by 
``impossible'' adversarial examples.   

\section{Auxiliary experiments}
\label{app:experiments}
In this section, we explore the space of possible parameters that we treated 
as fixed in the main paper. While this is not an exhaustive search, we 
hope to provide some intuition as to why we chose the parameters we did.
 
\subsection{Effect of $\lambda$ and $C$}
\begin{figure*}[t]
\begin{center}
\setlength{\tabcolsep}{0pt}
\includegraphics[align=c,width=2\columnwidth]{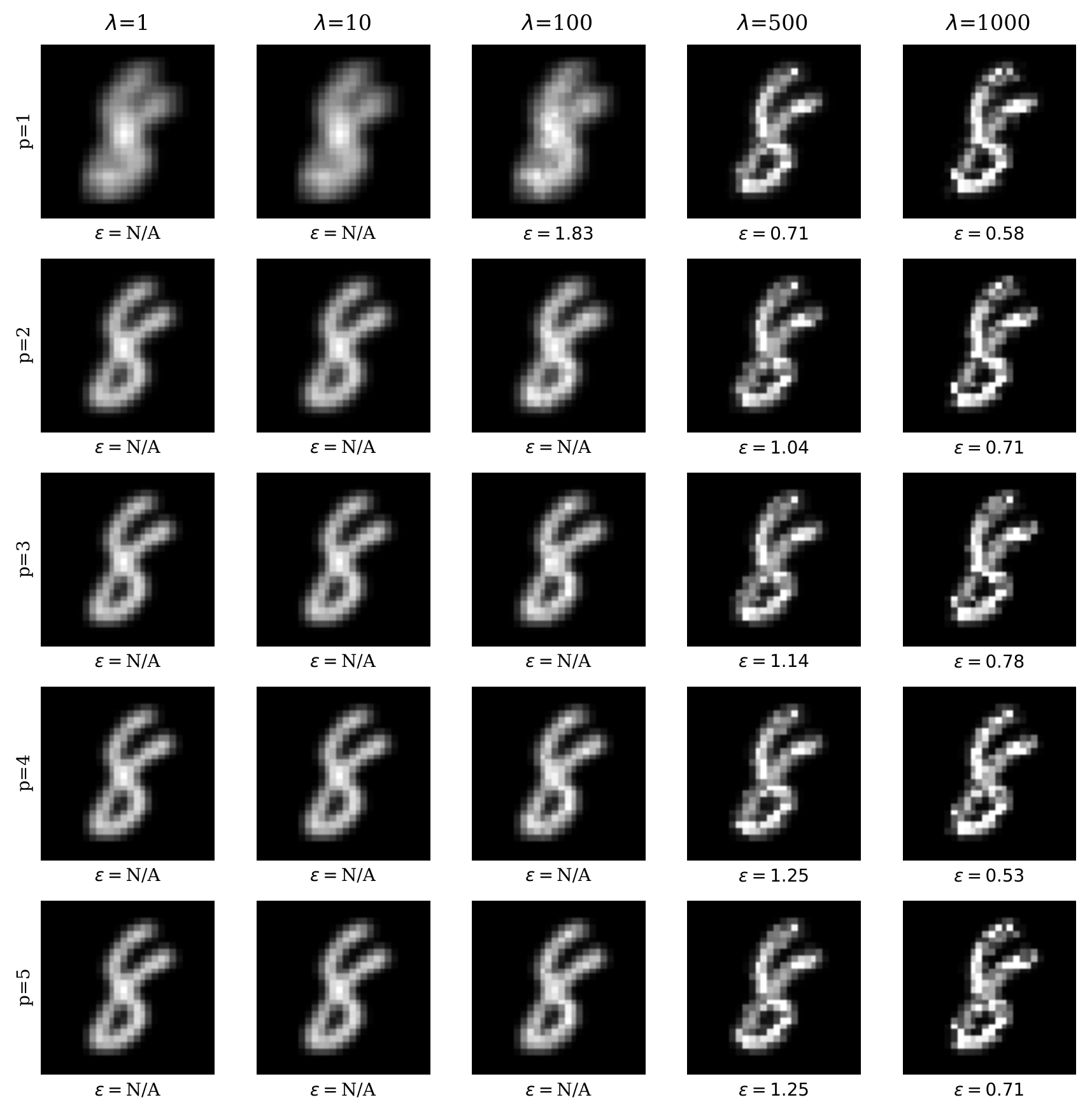} 

\vskip -0.1in
\caption{A plot of the adversarial examples generated with different $p$-Wasserstein metrics 
used for the cost matrix $C$ and different regularization parameters $\lambda$. Note that 
when regularization is low, the image becomes blurred, and it is harder to find adversarial examples. 
In contrast, changing $p$ does not seem to make any significant changes. 
}
\label{fig:lam_vs_p}
\end{center}
\vskip -0.2in
\end{figure*}

We first study the effect of $\lambda$ and the cost matrix $C$. First, note that 
$\lambda$ could be any positive value. Furthermore, 
note that to construct $C$ we used the 2-norm which reflects the 1-Wasserstein metric, 
but in theory we could use any $p$-Wasserstein metric, where 
the the cost of moving from pixel $(i,j)$ to $(k,l)$ is $\left(|i-j|^2 + |k-l|^2\right)^{p/2}$. Figure \ref{fig:lam_vs_p} 
shows the effects of $\lambda$ and $p$ on both the adversarial example and the radius 
at which it was found 
for varying values of $\lambda = [1, 10, 100, 500, 1000]$ and $p=[1,2,3,4,5]$.

We find that it is important to ensure that $\lambda$ is large enough, otherwise 
the projection of the image is excessively blurred. In addition to qualitative changes, 
smaller $\lambda$ seems to make it harder to find Wasserstein adversarial examples, 
making the $\epsilon$ radius go up as $\lambda$ gets smaller. In fact, for $\lambda = (1,10)$ and almost all of $\lambda = 100$, 
the blurring is so severe that no adversarial example can be found. 

In contrast, we find that increasing $p$ for the Wasserstein distance used in the cost matrix $C$ 
seems to make the images more ``blocky''. Specifically, as $p$ gets higher tested, more pixels seem to be moved in larger amounts. This seems to counteract the blurring observed for low $\lambda$ to some degree.  Naturally, the $\epsilon$ radius also grows since the overall cost of the transport plan has gone up. 

\begin{figure}[t]
\begin{center}
\setlength{\tabcolsep}{0pt}
\includegraphics[align=c,width=\columnwidth]{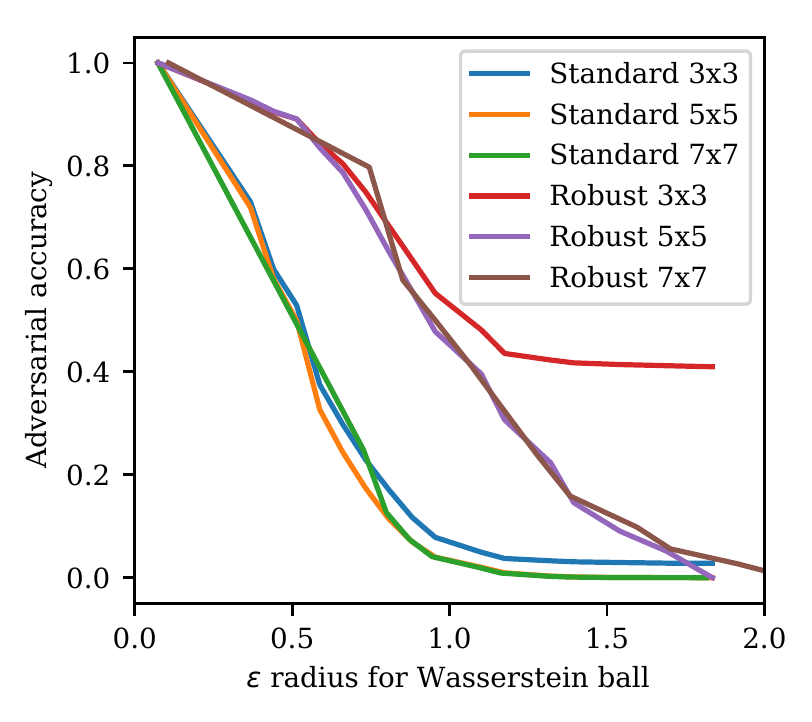} 

\vskip -0.1in
\caption{
Adversarial accuracy of a standard model and a model trained to be provably 
robust against $\ell_\infty$ attacks for different sizes of transport plans.
In most cases the size of the transport plan doesn't seem to matter, except 
for the $3\times 3$ local transport plan. 
In this case, the adversary isn't quite able to reach 0\% accuracy for the standard 
model, reaching 2.8\% for for $\epsilon=1.83$. The adversary is also 
unable to attack the robust MNIST model, bottoming out at 41\% adversarial accuracy at 
$\epsilon = 1.83$. 
}
\label{fig:transport_size}
\end{center}
\vskip -0.2in
\end{figure}

\subsection{Size of local transport plan}
In this section we explore the effects of different sized transport plans. In the main paper, 
we used a $5 \times 5$ local transport plan, but this could easily be something else, 
e.g. $3 \times 3$ or $7 \times 7$. We can see a comparison on the robustness of 
a standard and the $\ell_\infty$ robust model against these different sized 
transport plans in Figure \ref{fig:transport_size}, using $\lambda= 1000$. 
We observe that while $3\times 3$ transport plans have difficulty attacking the robust MNIST 
model, all other plan sizes seem to have similar performance. 


\section{Provable defense}
\label{app:provable}
In this section we show how a Sinkhorn-like algorithm can be derived 
for provable defenses, and that the resulting algorithm is actually 
just a simplified version of the Projected Sinkhorn iteration, 
which we call the Conjugate Sinkhorn iteration (since it solves 
the conjugate problem). 

\subsection{Conjugate Sinkhorn iteration}
By subtracting the same entropy term to the conjugate objective from Equation \eqref{eq:provable}, 
we can get a problem similar to that of projecting onto the Wasserstein ball.  
 \begin{equation}
\begin{split}
\minimize_{z\in\mathbb R^n_+, \Pi\in \mathbb{R}^{n\times n}_+} &\;\; -\lambda z^Ty + \sum_{ij}\Pi_{ij}\log(\Pi_{ij})\\
\subjectto \;\; & \Pi1 = x\\
& \Pi^T1 = z\\
& \langle \Pi,C\rangle \leq \epsilon. 
\end{split}
\end{equation}
where again we've multiplied the objective by $\lambda$ for convenience. 
Following the same framework as before, we introduce dual variables $(\alpha, \beta, \psi)$ 
where $\psi \geq 0$, to construct the Lagrangian as 
\begin{equation}
\begin{split}
&L(z, \Pi, \alpha, \beta, \psi) \\
= &-\lambda z^Ty + \sum_{ij}\Pi_{ij}\log(\Pi_{ij})+ \psi (\langle \Pi,C\rangle - \epsilon) \\
& + \alpha^T(x - \Pi1) + \beta^T(z - \Pi^T1). 
\end{split}
\end{equation}
Note that since all the terms with $\Pi_{ij}$ are the same, the corresponding KKT 
optimality condition for $\Pi_{ij}$ also remains the same. The only part that changes 
is the optimality condition for $z$, which becomes 
\begin{equation}
\beta = \lambda y
\end{equation}
Plugging the optimality conditions into the Lagrangian, we get the following dual problem: 
\begin{equation}
\begin{split}
& L(z^*,\Pi^*, \alpha, \beta, \psi) \\
=& - \psi \epsilon + \alpha^Tx \\
& - \sum_{ij}\exp(\alpha_i)\exp( - \psi C_{ij} - 1)\exp( \beta_j)\\
=& g(\alpha, \psi)
\end{split}
\end{equation}
Finally, if we minimize this with respect to $\alpha$ and $\psi$ we get exactly 
the same update steps as the Projected Sinkhorn iteration. Consequently, the Conjugate Sinkhorn iteration 
is identical to the Projected Sinkhorn iteration except that 
we replace the $\beta$ step with the fixed value $\beta = \lambda y$.


\end{document}